\documentclass[letterpaper, 10 pt, journal, twoside]{ieeetran}
\usepackage{amsmath,amsfonts}
\usepackage{algorithmic}
\usepackage{algorithm}
\usepackage{array}
\usepackage[caption=false,font=normalsize,labelfont=sf,textfont=sf]{subfig}
\usepackage{textcomp}
\usepackage{stfloats}
\usepackage{url}
\usepackage{verbatim}
\usepackage{graphicx}
\usepackage{cite}
\usepackage{xspace}

\usepackage{siunitx}
\usepackage{booktabs}
\usepackage{newunicodechar}
\newunicodechar{̌}{\v{ }}
\usepackage{multirow}
\newcommand{\best}[1]{\textbf{#1}}
\usepackage{makecell}
\usepackage{xcolor}
\usepackage[colorlinks=true, urlcolor=blue!60!black, linkcolor=black, citecolor=black]{hyperref}

\begin{document}

\newcommand{\name}{REACT3D}
\newcommand{\myparagraph}[1]{\noindent\textbf{#1.}\;}
\newcommand{\myparagraphbaselines}[1]{\noindent\textit{#1.}\;}

\title{REACT3D:\\Recovering Articulations for Interactive Physical 3D Scenes}
\author{Zhao Huang$^{1}$, Boyang Sun$^{1}$, Alexandros Delitzas$^{1,2}$, Jiaqi Chen$^{1}$, and Marc Pollefeys$^{1,3}$%
\thanks{Manuscript received October 7, 2025; Revised January 15, 2026; Accepted February 24, 2026.}
\thanks{This paper was recommended for publication by Editor Abhinav Valada upon evaluation of the Associate Editor and Reviewers’ comments.}
\thanks{This work was supported by the
Swiss National Science Foundation Advanced Grant
216260: “Beyond Frozen Worlds: Capturing Functional
3D Digital Twins from the Real World”, and the European Union’s Horizon Europe research and innovation programme under grant agreement number 101214398 (ELLIOT). Alexandros Delitzas is supported
by the Max Planck ETH Center for Learning Systems
(CLS).}


\thanks{$^{1}$All authors are with ETH Zurich, 8092 Zurich, Switzerland {\footnotesize \url{zhahuang@ethz.ch}}; {\footnotesize \url{boyang.sun@inf.ethz.ch}}; {\footnotesize \url{alexandros.delitzas@inf.ethz.ch}}; {\footnotesize \url{jiaqi.chen@inf.ethz.ch}}; {\footnotesize \url{marc.pollefeys@inf.ethz.ch}}}
\thanks{$^{2}$Alexandros Delitzas is also with Max Planck Institute for Informatics, 66123 Saarbruecken, Germany {\footnotesize \url{alexandros.delitzas@mpi-inf.mpg.de}}}

\thanks{$^{3}$Marc Pollefeys is also with Microsoft Spatial AI Lab, 8038 Zurich, Switzerland {\footnotesize \url{mapoll@microsoft.com}}}

\thanks{Digital Object Identifier (DOI):  10.1109/LRA.2026.3674028}
}

\markboth{IEEE Robotics and Automation Letters. Preprint Version. Accepted February, 2026}%
{Huang \MakeLowercase{\textit{et al.}}: REACT3D: Recovering Articulations for Interactive Physical 3D Scenes}

\IEEEaftertitletext{%
\begin{center}
\vspace{-2.5em}
{\small Project page: \underline{\href{https://react3d.github.io}{react3d.github.io}}}
\end{center}
\vspace{0.6em}
}
\maketitle

\begin{abstract}
Interactive 3D scenes are increasingly vital for embodied intelligence, yet existing datasets remain limited due to the labor-intensive process of annotating part segmentation, kinematic types, and motion trajectories. We present \name{}, a scalable zero-shot framework that converts static 3D scenes into simulation-ready interactive replicas with consistent geometry, enabling direct use in diverse downstream tasks. Our contributions include: (i) openable-object detection and segmentation to extract candidate movable parts from static scenes, (ii) articulation estimation that infers joint types and motion parameters, (iii) hidden-geometry completion followed by interactive object assembly, and (iv) interactive scene integration in widely supported formats to ensure compatibility with standard simulation platforms. We achieve state-of-the-art performance on detection/segmentation and articulation metrics across diverse indoor scenes, demonstrating the effectiveness of our framework and providing a practical foundation for scalable interactive scene generation, thereby lowering the barrier to large-scale research on articulated scene understanding.
\end{abstract}


\section{Introduction}

\IEEEPARstart{H}{igh-fidelity}, interactive 3D assets are critical for a wide range of applications. They enable immersive experiences in virtual and mixed reality, support realistic content creation for gaming and film production, and facilitate the development of autonomous robotic systems by serving as training environments for navigation and manipulation tasks~\cite{puig2023habitat3}. These applications demand large-scale 3D datasets that offer both photorealistic rendering and physically plausible interactions, such as object picking, placing, or articulating structures like doors and drawers. Automating the generation of such datasets is essential to scale up these efforts effectively.

Significant effort has been dedicated to the automatic generation of 3D assets. Early works primarily focused on the creation of \textit{static} 3D assets, often converting real-world scans or image sequences into accurate 3D representations such as meshes or point clouds~\cite{yeshwanthliu2023scannetpp}. These efforts have led to the release of several high-quality static 3D datasets. More recently, the rapid advancement of generative models has enabled the synthesis of large-scale static scenes with diverse modalities and formats~\cite{zhai2024echoscene}. As a result, the generation of static 3D scenes, whether through real-world reconstruction or generative modeling, has become increasingly mature.


\begin{figure}[!t]
  \centering
  \includegraphics[width=\columnwidth]{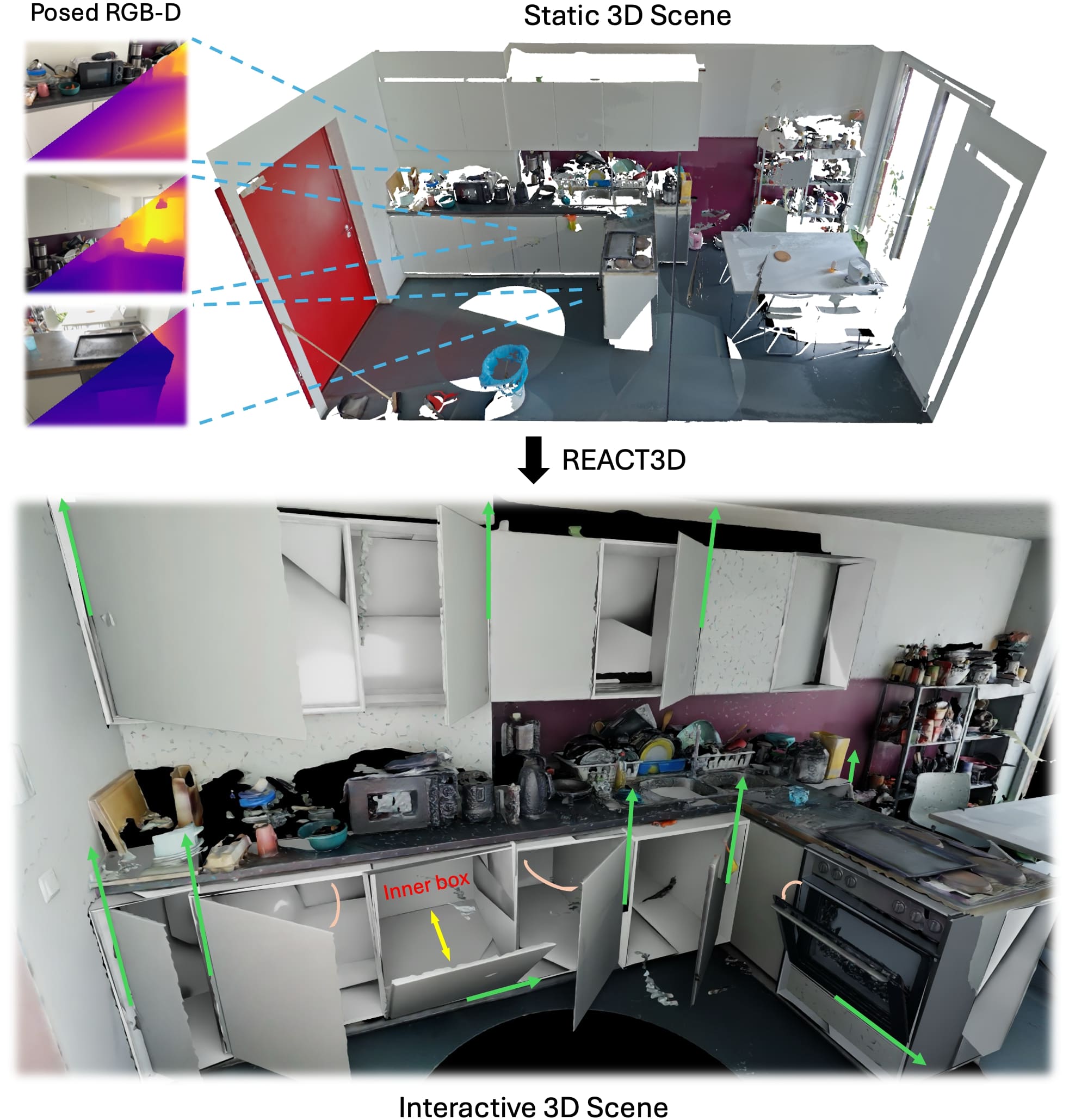}
  \caption{\textbf{\name{}} transforms static 3D scenes into interactive scenes in a zero-shot manner. The generated interactive scenes are spatially aligned with the static input and preserve the original geometry and appearance. Our results are readily compatible with multiple simulation platforms, supporting diverse downstream tasks such as robotic perception, interaction, and embodied intelligence.}
  \vspace{-0.3em}
  \label{fig:teaser}
\end{figure}

In parallel, the research community has begun to explore the generation of \textit{dynamic} scenes, either by reconstructing real-world interaction sequences~\cite{xia2025drawerdigitalreconstructionarticulation} or by manually designing 3D environments with humans in the loop. However, despite promising progress, the scalability and quality of interactive scene generation still lag behind those of static scene generation.
\IEEEpubidadjcol

In this work, we are motivated by the growing need for large-scale, interactive 3D environments to support embodied AI research. We propose a novel, application-driven framework for converting high-quality static scenes into physics-aware, interactive environments (Fig.~\ref{fig:teaser}). Our system leverages recent advances in vision foundation models to enable zero-shot transfer from standard 3D scene formats (e.g., \texttt{.ply}, \texttt{.glb}) to simulation-ready formats such as URDF and USD. These converted scenes support physical simulation and enable robotic agents to navigate and interact with articulated objects. Our method provides a reliable and efficient solution for obtaining large-scale, physics-aware 3D assets, without requiring additional real-world data collection or computationally intensive generation. By combining the abundance of static 3D datasets with the strengths of vision-based perception models, our approach offers a scalable and generalizable pathway toward dynamic scene generation. The output of our pipeline can be seamlessly integrated into a wide range of renderers, simulators, and ecosystems, such as ROS, Isaac Sim, PyBullet, and Open3D, enabling flexible deployment across diverse downstream tasks. In summary, our key contributions are:

\begin{itemize}
    \item We present REACT3D, an efficient, automated workflow that leverages vision foundation models and vision-language models (VLMs) to recover object articulations from static 3D scenes and generate physically-enabled 3D scenes.
    \item We provide a comprehensive evaluation of our pipeline against baseline methods.
    \item We demonstrate the utility of our output assets across various platforms, including renderers, physics simulators, and the ROS ecosystem.
\end{itemize}

\begin{figure*}[t]
  \centering
  \includegraphics[width=0.98\textwidth]{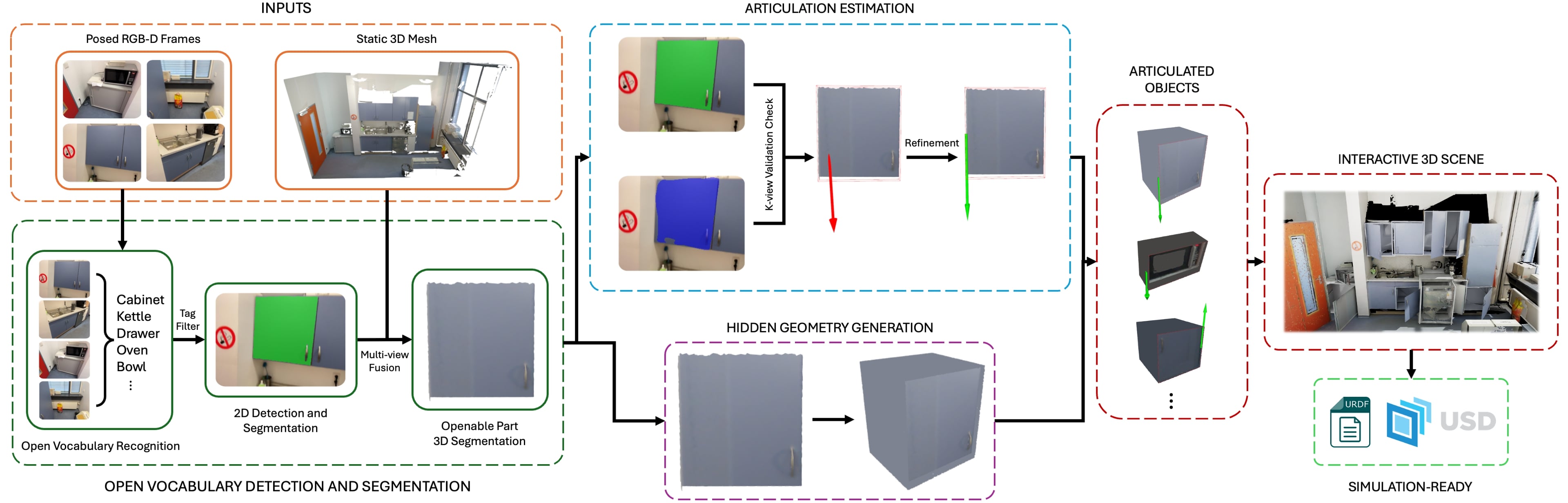}
  \caption{Overview of \name{}. Given a static 3D scene, our method first applies open-vocabulary detection to identify openable objects and segmentation to extract their movable parts. We then estimate articulations and generate hidden geometry to obtain interactive objects. Finally, they are integrated with the static background to produce a simulation-ready interactive scene.}
  \label{fig:overview}
  \vspace{-0.5em}
\end{figure*}

\section{Related Work}
\subsection{Articulation estimation}

Articulated object motion estimation is a long-standing problem in computer vision and robotics~\cite{mobility_trees, jiang2022opd, mao2022multiscan}. Numerous works have explored predicting openable object parts and their motion parameters~\cite{jiang2022opd, sun2023opdmulti, qian2023understanding}.
OPD~\cite{jiang2022opd} addresses this task by extending Mask R-CNN~\cite{maskrcnn} to detect articulated parts and estimate their motions from single-view object images.
OPDMulti~\cite{sun2023opdmulti} further generalizes OPD to handle real-world scenes containing multiple objects. In addition, 3DOI~\cite{qian2023understanding} proposes a transformer-based model to estimate the physical properties and affordance of objects using a single image and 2D query points as input.

Another line of work~\cite{mao2022multiscan, delitzas2024scenefun3d, halacheva2024articulate3d} focuses on predicting articulation directly on scene-level point clouds.
SceneFun3D~\cite{delitzas2024scenefun3d} introduces a large-scale dataset aimed at understanding object functionalities and affordances in real-world 3D environments, along with baseline models for predicting motion parameters associated with interactions involving functional elements (e.g., opening a drawer by pulling its handle).
Articulate3D~\cite{halacheva2024articulate3d} builds upon ScanNet++~\cite{yeshwanthliu2023scannetpp} annotated scenes to provide object-part hierarchy annotations and articulation parameters.
However, scene-level point clouds often lack the fine-grained details of 2D images, and their annotations remain sparse. In this work, we propose an open-vocabulary approach to segment and articulate openable parts by leveraging off-the-shelf image-based articulation estimators together with segmentation foundation models.

\subsection{Articulated scenes from static observations}
Early efforts on building interactive 3D scenes often required observing objects in multiple states or under user interaction. For instance, Ditto~\cite{jiang2022ditto} reconstructs part geometry and joints by comparing 3D observations before and after human interaction. Similarly, PARIS~\cite{jiayi2023paris} uses two sets of multi-view images captured under different articulation states to jointly recover part-level shape and motion parameters. Weng \emph{et al.}~\cite{weng2024neural} likewise leverage two RGB-D scans of an object at different articulations to learn a neural implicit twin with accurate kinematics. While these methods achieve faithful motion recovery, their reliance on paired observations or physical trials limits scalability. Some works have explored video-based static capture: e.g., ArtGS~\cite{liu2025artgsbuildinginteractablereplicas} employs 3D Gaussian splatting on casually captured video frames to reconstruct photorealistic objects with movable parts, but still benefits from careful segmentation or coarse templates to handle occluded regions. Recently,  FunRec~\cite{Delitzas2026FunRec} and iTACO~\cite{peng2025itaco} further explore reconstructing articulated scenes directly from interaction videos. Overall, approaches that demand explicit state changes incur significant manual interaction overhead and scale poorly to large-scale scene conversion.

Recent research instead focuses on recovering articulated scenes directly from single-state observations using learned priors and foundation models. At the object level, Real2Code~\cite{mandi2024real2codereconstructarticulatedobjects} proposes to generate programmatic code describing parts and joints given RGB images, using VLM to enforce structural consistency. Articulate-Anything~\cite{le2024articulate} retrieves a similar 3D model from a shape library and infers plausible joints, enabling articulation for arbitrary objects but at the cost of fidelity to the actual geometry. Other methods tackle full scenes from minimal input: Digital Cousins~\cite{dai2024acdc} and URDFormer~\cite{chen2024urdformerpipelineconstructingarticulated} generate an interactive scene from a single RGB image by leveraging prior knowledge. However, single-image approaches often yield inconsistent or non-faithful results, since a single view provides only partial evidence. More comprehensive pipelines integrate multi-view reconstruction and segmentation cues. DRAWER~\cite{xia2025drawerdigitalreconstructionarticulation} reconstructs scenes from posed RGB frames using foundation models for part segmentation and articulation estimation, together with neural fields for appearance, achieving state-of-the-art realism and accurate articulations. These advances demonstrate a clear trend toward zero-shot articulated scene generation from static inputs, but challenges remain in balancing fidelity, physical accuracy, and scalability.

\section{Method}
\label{sec:method}

\subsection{Preliminaries}
\myparagraph{Problem formulation}
Given a static 3D scene representation \(\mathbf{S}\), our goal is to construct an interactive digital twin \(\hat{\mathbf{S}}\) that augments the original geometry with articulated objects while preserving the static background (Fig.~\ref{fig:overview}). The input \(\mathbf{S}\) consists of a point cloud or a mesh \(\mathbf{P}\) with per-vertex colors, captured by a registered RGB-D sensor array, together with a set of calibrated frames \(\{(\mathbf{I}_i, \mathbf{D}_i, \mathbf{T}_i, \mathbf{K}_i)\}_{i=1}^N\), where \(\mathbf{I}_i\) is an RGB image, \(\mathbf{D}_i\) its depth map, \(\mathbf{T}_i \in SE(3)\) the extrinsic pose, and \(\mathbf{K}_i \in \mathbb{R}^{3\times 3}\) the intrinsic matrix. The output \(\hat{\mathbf{S}}\) is decomposed as
\[
\hat{\mathbf{S}} = f(\mathbf{S}) \;=\; \big(\,\mathbf{P}_{bg},\ \{\mathbf{O}_j\}_{j=1}^M\big),
\quad
\mathbf{O}_j = \big(\mathbf{p}_j,\, \phi_j,\, \mathbf{b}_j\big),
\]
where \(\mathbf{P}_{bg}\) denotes the static background and each interactive object \(\mathbf{O}_j\) is defined by its movable part \(\mathbf{p}_j\), articulation parameters \(\phi_j\), and completed hidden geometry \(\mathbf{b}_j\). The articulation is parameterized as
\[
\phi_j = (t_j, \mathbf{o}_j, \mathbf{a}_j, \rho_j),
\]
with joint type \(t_j \in \{\mathrm{prismatic},\, \mathrm{revolute}\}\), origin \(\mathbf{o}_j \in \mathbb{R}^3\), axis \(\mathbf{a}_j \in \mathbb{R}^3\), and motion limits \(\rho_j\).

\myparagraph{Openable object}
We define an openable object as one that contains a movable part actuated by either a prismatic or a revolute joint~\cite{li2020categorylevelarticulatedobjectpose}. For a prismatic joint, the part translates along a fixed axis \(\mathbf{a}\) with displacement \(s \in [0, \rho]\), while for a revolute joint, it rotates about a unit axis \(\mathbf{a}\) through origin \(\mathbf{o}\) with angle \(\theta \in [0, \rho]\). 
The motion of any point \(\mathbf{x}\) belonging to the movable part in its closed configuration is
\[
\mathbf{x}' \;=\;
\begin{cases}
\mathbf{x} + s\,\mathbf{a}, & \text{prismatic}, \\[6pt]
R(\mathbf{a},\theta)(\mathbf{x}-\mathbf{o}) + \mathbf{o}, & \text{revolute},
\end{cases}
\]
where \(R(\mathbf{a},\theta)=\exp([\mathbf{a}]_\times \theta)\) denotes the axis-angle rotation and \([\mathbf{a}]_\times\) is the skew-symmetric matrix of \(\mathbf{a}\). This yields a compact representation of each movable part's motion, parameterized by \((t,\mathbf{o},\mathbf{a},\rho)\), which serves as the basis for kinematic prediction.

\subsection{Openable object detection and segmentation}

Interactive scene understanding requires semantic knowledge of object affordances beyond static 3D geometry. The detection module is utilized to identify candidate interactive instances, while part segmentation delineates movable components from the structure to obtain aligned meshes. This decomposes the static scene into instance-level proposals for downstream articulation estimation, determining the extent of interactivity in the resulting scene.

\myparagraph{Open-vocabulary 2D detection and segmentation}
Indoor scenes include diverse openable objects not captured by standard labels, motivating an open-vocabulary detection strategy to maximize interactive coverage. We therefore first process the RGB frames \(\{\mathbf{I}_i\}_{i=1}^N\) of the static scene to recognize semantic tags of objects present in the scene using RAM++~\cite{zhang2023recognizeanythingstrongimage}. Let \(\mathcal{T} = \bigcup_{i=1}^N \texttt{RAM++}(\mathbf{I}_i)\) denote the aggregated tag set. We then apply a VLM-based filtering method built on LLaVA~\cite{liu2023llava} to retain only tags that satisfy our definition of openable objects, yielding \(\mathcal{T}_{\mathrm{open}} = \{\, t \in \mathcal{T} \mid f_{\mathrm{VLM}}(t)=1 \,\}\). Finally, Grounded~SAM~\cite{ren2024grounded} is applied with prompts \(\mathcal{T}_{\mathrm{open}}\) on each frame \(\mathbf{I}_i\) to localize and segment movable parts, producing a set of 2D instance masks \(\{\mathbf{m}_{i,k}\}_{k=1}^{n_i}\), where \(n_i\) denotes the number of detected movable parts in frame \(\mathbf{I}_i\), which are post-processed with hole filling to close interior gaps, as Grounded~SAM can miss door handles. Unlike prior approaches restricted to predefined labels, our open-vocabulary method mitigates label bias and yields better coverage of long-tail openable objects.

\begin{figure}[tbp]
  \centering

  \begin{minipage}[b]{0.32\linewidth}
    \centering
    \scriptsize\textbf{Input}\\[0.1em]
    \includegraphics[width=\linewidth]{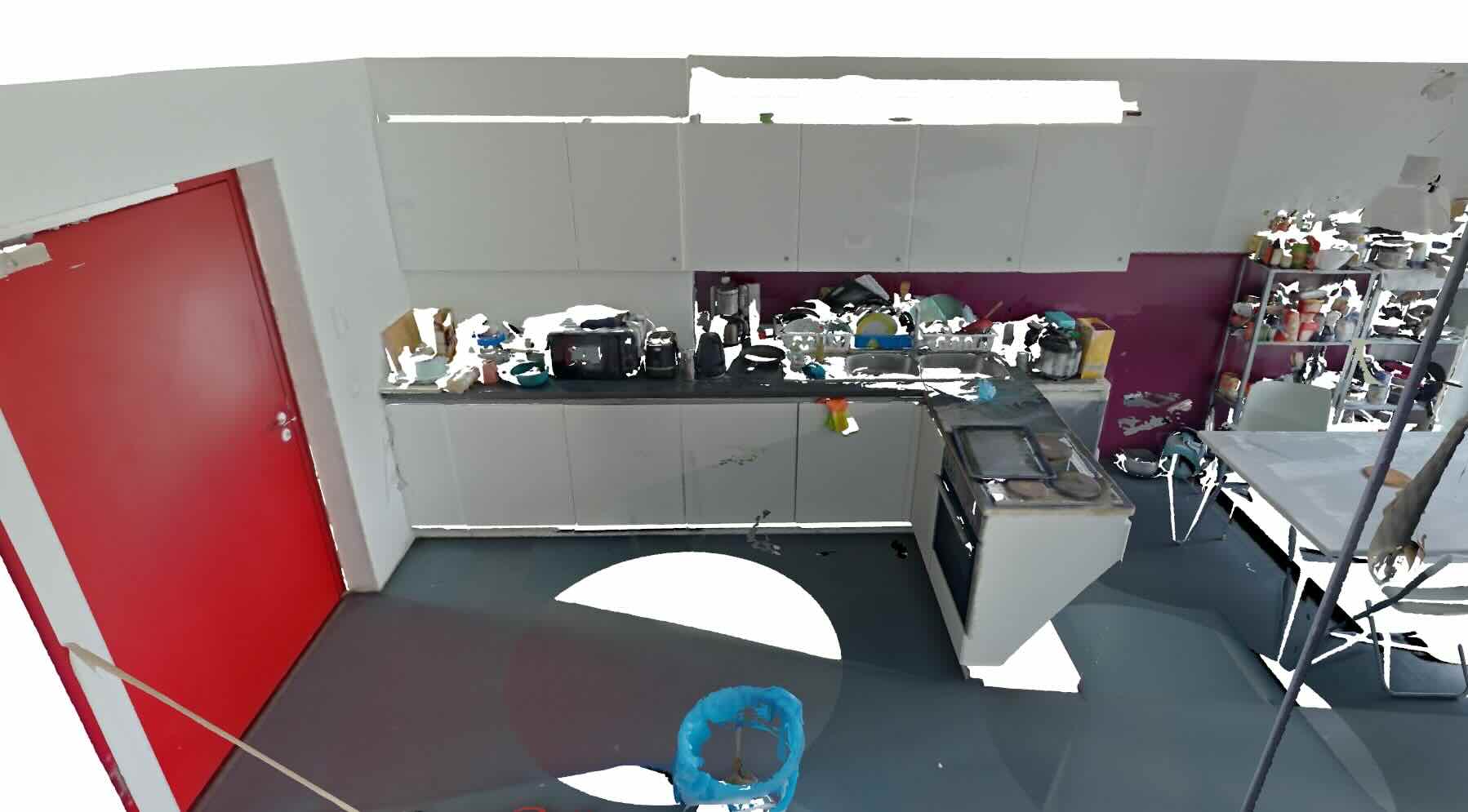}
  \end{minipage}
  \hfill
  \begin{minipage}[b]{0.32\linewidth}
    \centering
    \scriptsize\textbf{DRAWER}\\[0.1em]
    \includegraphics[width=\linewidth]{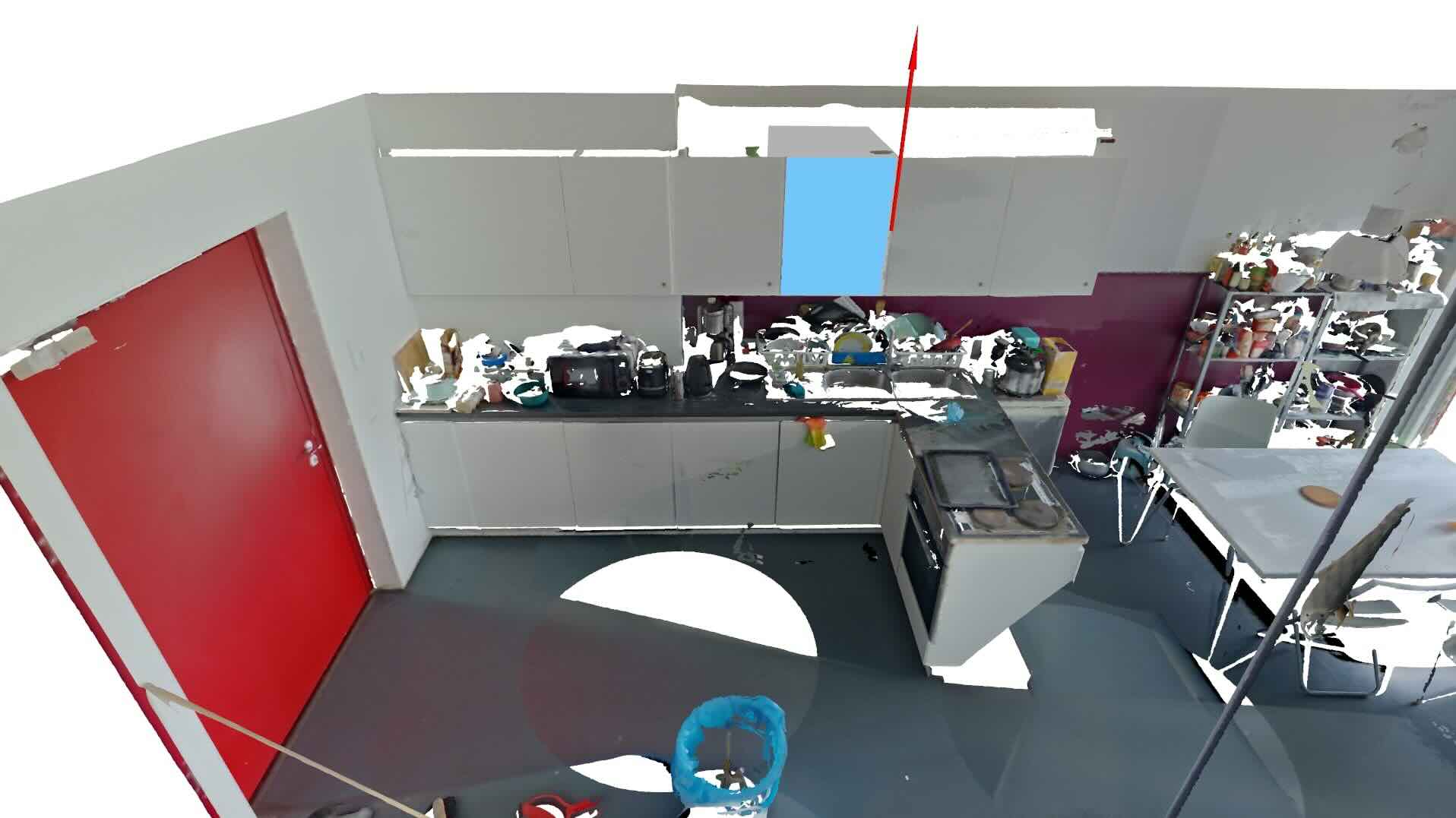}
  \end{minipage}%
  \hfill
  \begin{minipage}[b]{0.32\linewidth}
    \centering
    \scriptsize\textbf{\name{}}\\[0.1em]
    \includegraphics[width=\linewidth]{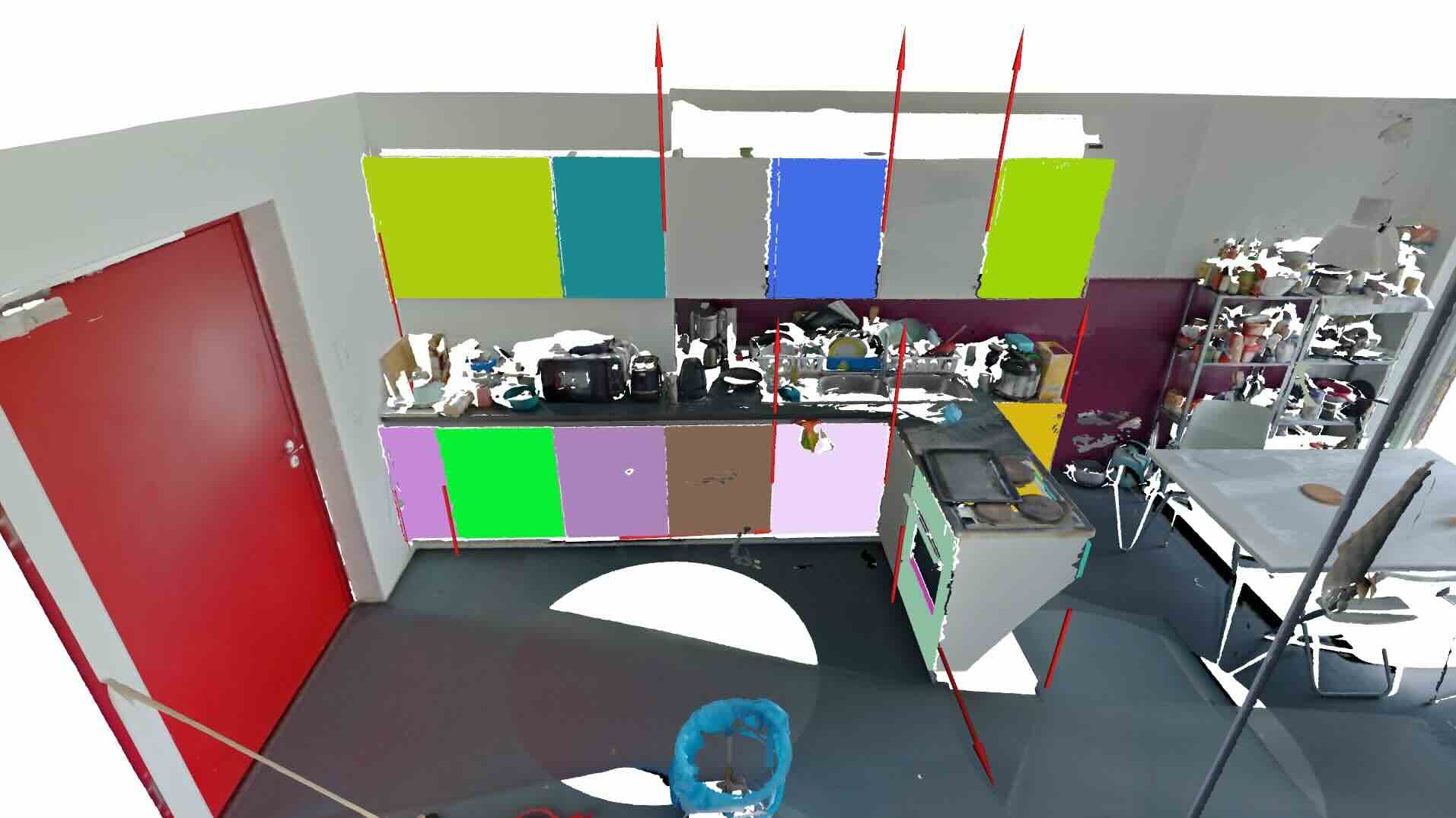}
  \end{minipage}

  \vspace{0.5em}

  \begin{minipage}[b]{0.32\linewidth}
    \centering
    \includegraphics[width=\linewidth]{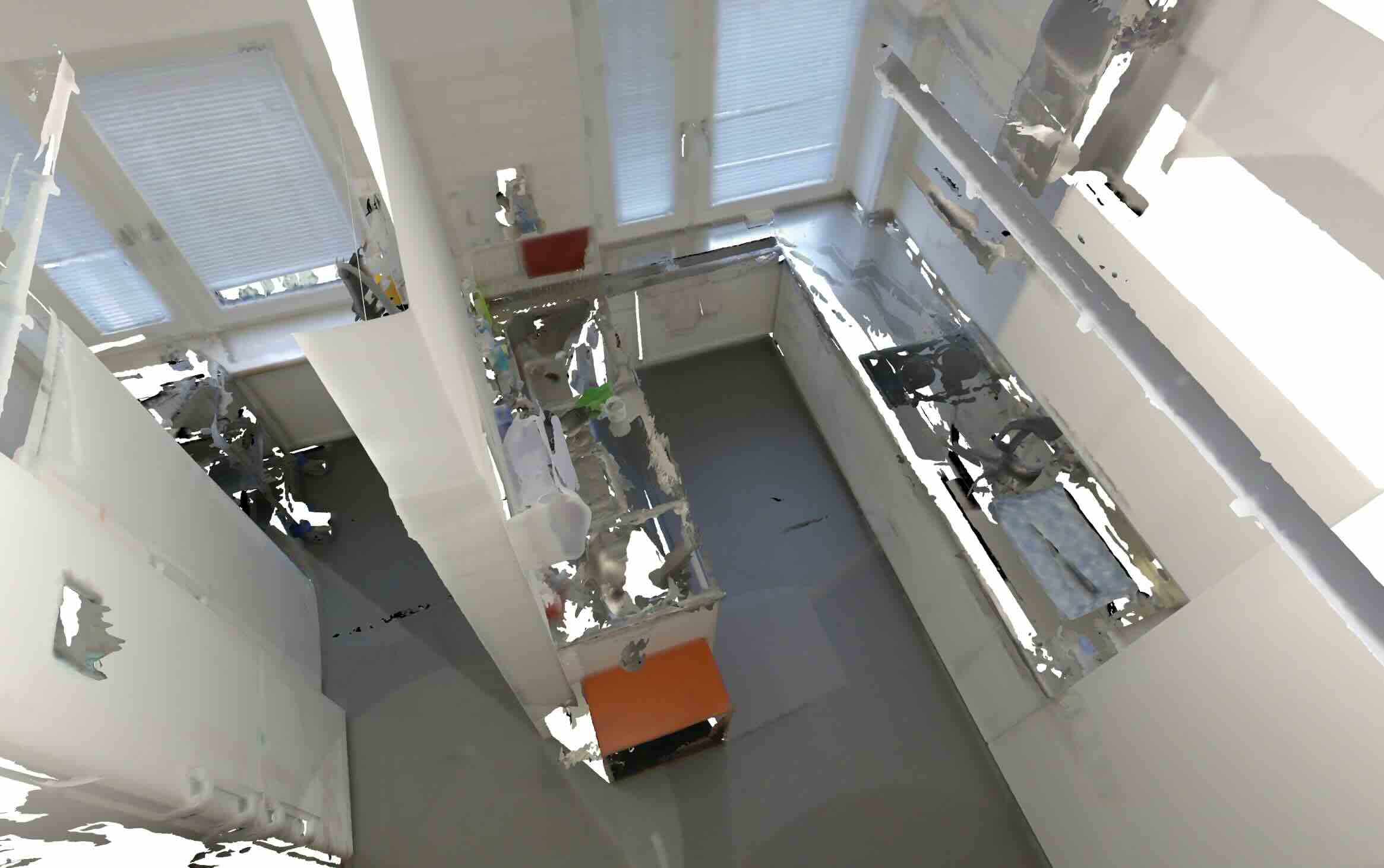}
  \end{minipage}%
  \hfill
  \begin{minipage}[b]{0.32\linewidth}
    \centering
    \includegraphics[width=\linewidth]{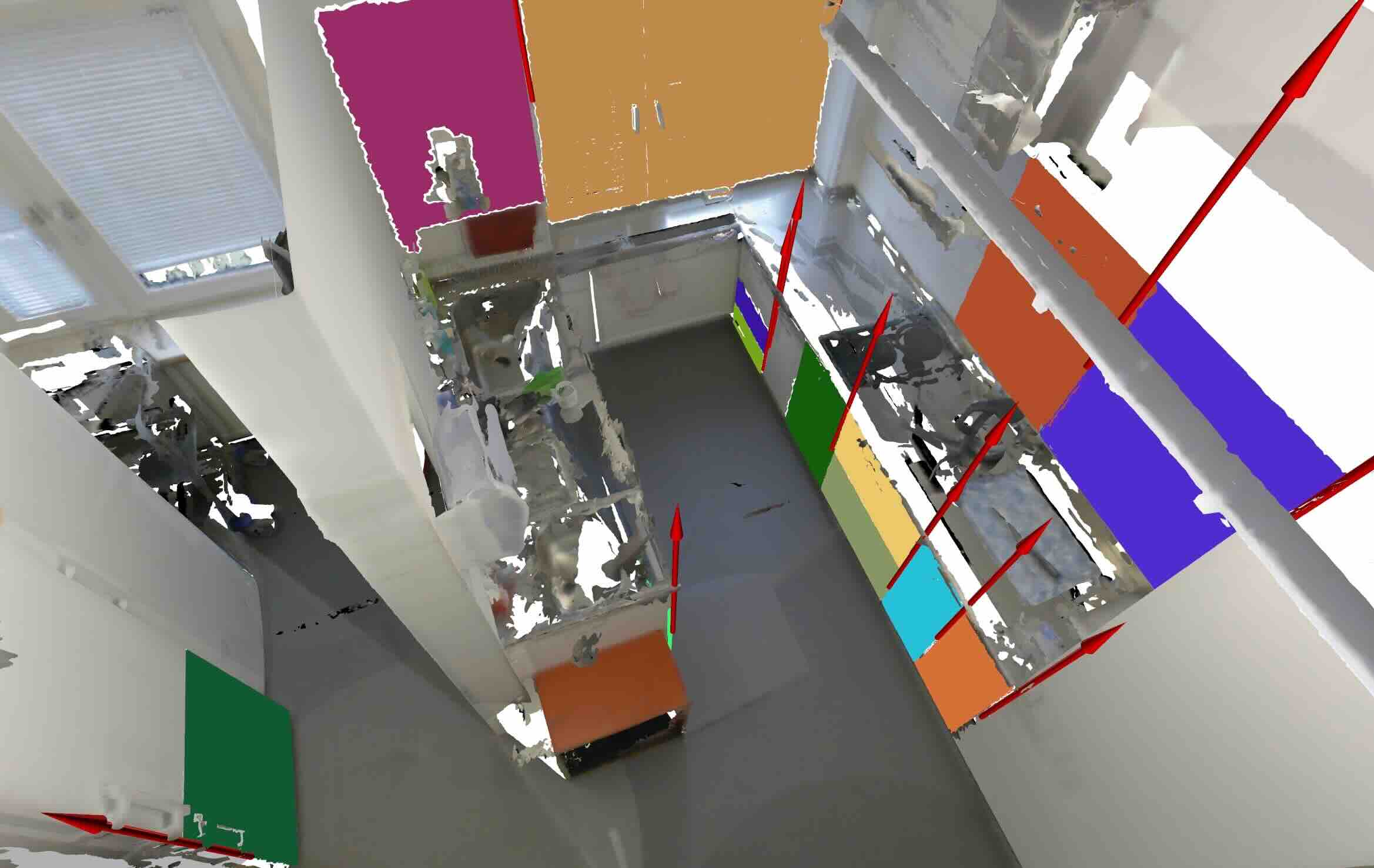}
  \end{minipage}%
  \hfill
  \begin{minipage}[b]{0.32\linewidth}
    \centering
    \includegraphics[width=\linewidth]{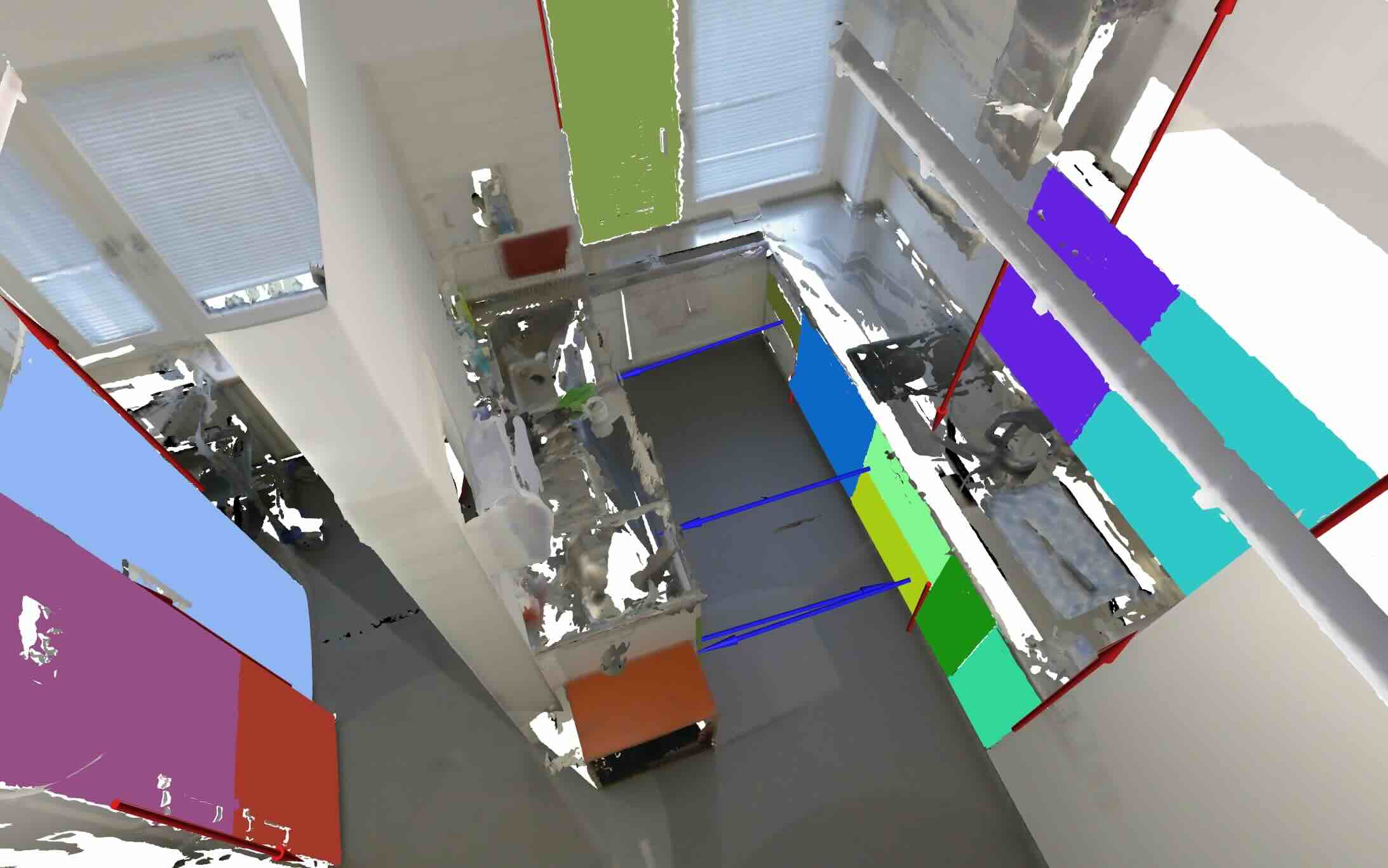}
  \end{minipage}


  \caption{Qualitative comparison. \textcolor{red}{Red} and \textcolor{blue}{blue} arrows denote revolute and prismatic joints. Our method outperforms DRAWER on handle-less objects (top) and correctly classifies prismatic drawers that DRAWER mistakes for revolute joints (bottom).}
  \label{fig:part_arti_comparison}
\end{figure}

\myparagraph{2D-to-3D segmentation via multi-view fusion}
Zero-shot 3D instance segmentation methods~\cite{Schult23ICRA} operating on point clouds or meshes rarely yield reliable \emph{part}-level segmentation. We therefore adopt a 2D-to-3D multi-view fusion paradigm, similar to DRAWER, to lift robust 2D masks into 3D. For each frame, we construct the model-view-projection matrix from \(\mathbf{T}_i, \mathbf{K}_i\) to recover the camera view. The scene mesh is then rasterized under each view, and the front-most faces covered by each mask are selected as its 3D projection. To recover the full structure of each object, as they are not always fully visible from a single view, we fuse the per-mask projections by building a face graph and applying Louvain community detection~\cite{do2024improvementlouvainalgorithmusing}. Unlike DRAWER, which retains only one mask above the Intersection over Union (IoU) threshold for each fusion result, our approach ranks the masks by IoU and preserves the top-\(k\) views that exceed the threshold, providing multiple perspectives that improve the robustness of subsequent articulation estimation. In addition, we replace its VLM-based filtering of part masks, which discards candidates without visible handles, with an alternative validation step at the articulation estimation stage (Sec.~\ref{subsec:articulation_estimation}). This design allows the detection of openable objects even when handles are absent or visually nonsalient, as shown by the comparison in Fig.~\ref{fig:part_arti_comparison}. We then re-segment each object by invoking SAM with point prompts seeded from previously fused 3D proposals.

Another challenge in 2D-to-3D lifting is that segmentation often produces the whole object rather than its movable part. To address this, we refine each segmented 3D mesh by applying RANSAC~\cite{10.1145/358669.358692} to estimate an approximately vertical, finite-thickness plane and retain the estimate with the largest contour area. Naive planar clipping, however, may fail to capture protruding handles. To preserve such structures while discarding static body regions, let $\mathbf{n}$ denote the normal of the estimated plane and $\mathbf{v}_i$ the viewing direction derived from the camera pose $\mathbf{T}_i$ of the frame that produced the segmentation. We choose the inward-pointing normal such that \(\langle \mathbf{n}, \mathbf{v}_i \rangle > 0\), and discard the geometry behind the plane along this direction. This produces high-quality 3D movable-part meshes $M^{\mathrm{obj}}_{j}=(V_j, F_j)$, where \(V_j\) and \(F_j\) denote the vertex and face sets of the segmented mesh, together with the top-$k$ RGB views and their associated masks $\{\mathbf{m}_{j,i}\}_{i=1}^k$. The resulting part meshes are spatially aligned with the input static scene and provide the foundation for generating interactive objects.

\subsection{Articulation estimation}
\label{subsec:articulation_estimation}
To enable meaningful manipulation, each movable part must be endowed with kinematic parameters that define its motion. Accordingly, for every movable part, we estimate an articulation tuple \(\phi(m)=\big(t,\mathbf{o},\mathbf{a},\rho\big)\). In our framework, we leverage OPDMulti~\cite{jiang2022opd, sun2023opdmulti} to obtain initial articulation estimates and corresponding movable part masks.

\myparagraph{Candidate filtering}
Since OPDMulti can detect multiple movable parts from a single RGB frame, it offers the opportunity to further validate the candidate openable objects obtained from the segmentation stage. For each segmented 3D mesh, we utilize the pre-saved top-\(k\) best-view images and their SAM-generated masks \(M_{\mathrm{SAM}}=\{\mathbf{m}_{i}\}_{i=1}^k\). We then run OPDMulti on these views and compare its predicted movable-part masks \(M_{\mathrm{OPDM}}\) against \(M_{\mathrm{SAM}}\). A candidate is considered valid if the IoU between the two masks exceeds a threshold in at least one of the top-\(k\) views; otherwise, the 3D mesh is discarded as an invalid case.

\myparagraph{Articulation refinement}
While OPDMulti provides a reasonable initial articulation, direct application to interactive objects may fail to reproduce realistic joint behavior. To recover faithful interactions, we refine the initial estimates for each validated movable part by computing an oriented bounding box (OBB) of the part mesh with COMPAS~\cite{compas-dev}. The OBB is denoted \((\mathbf{U}, \mathbf{c}, \mathbf{s})\), where \(\mathbf{U}=[\mathbf{u}_1\,\mathbf{u}_2\,\mathbf{u}_3]\) is a set of orthogonal edges, \(\mathbf{c}\) is the center, and 
$\mathbf{s} = (s_1, s_2, s_3)$ are the side lengths with $s_1 \geq s_2 \geq s_3$. Since openable parts are typically plate-like, we define the front face \(P_{\mathrm{front}}\) as the OBB face spanned by the two longest edges \(\mathbf{u}_1\) and \(\mathbf{u}_2\) with unit normal \(\mathbf{n}_{\mathrm{front}} = \mathbf{u}_1 \times \mathbf{u}_2\). Let the initial articulation be \(\phi(m)=(t,\mathbf{o},\mathbf{a},\rho)\). For prismatic joints, we update only the joint axis direction by aligning it with the front-face normal,
\[
\mathbf{a}' \;=\; \operatorname{sign}\!\big(\langle \mathbf{a}, \mathbf{n}_{\mathrm{front}}\rangle\big)\,\mathbf{n}_{\mathrm{front}},\qquad
\phi'(m)=(t,\mathbf{o},\mathbf{a}',\rho),
\]
which preserves the origin \(\mathbf{o}\) while enforcing translation orthogonal to \(P_{\mathrm{front}}\). For revolute joints, we first select the in-plane principal direction most consistent with the initial axis,
\[
\mathbf{l}_s \;=\; \arg\max_{\mathbf{v}\in\{\mathbf{u}_1,\mathbf{u}_2\}} \, \big|\langle \mathbf{a}, \mathbf{v}\rangle\big|,\qquad
\mathbf{a}' \;=\; \operatorname{sign}\!\big(\langle \mathbf{a}, \mathbf{l}_s\rangle\big)\,\mathbf{l}_s.
\]
Next, we place the origin by comparing the joint axis line defined by the initial axis
\(
L(\mathbf{o},\mathbf{a})=\{\mathbf{o}+s\,\mathbf{a}\mid s\in\mathbb{R}\}
\)
to the two OBB front-face edges that are parallel to \(\mathbf{l}_s\). We compute the shortest distance from \(L\) to each of these edges, select the closer one, and take its midpoint. This point is then projected along \(\mathbf{n}_{\mathrm{front}}\) onto the mid-surface equidistant between \(P_{\mathrm{front}}\) and its parallel opposite OBB face \(P_{\mathrm{back}}\), where the projected point serves as the refined joint origin \(\mathbf{o}'\). This procedure anchors the origin to the mid-surface edge most consistent with the initial axis while ensuring the most appropriate location within the part's thickness. Finally, we update
\[
\phi'(m)=(t,\mathbf{o}',\mathbf{a}',\rho).
\]
Intuitively, the prismatic axis is made orthogonal to the part face, whereas the revolute axis is aligned with the dominant in-plane edge and centered within the part thickness; in both cases the axis sign is chosen to minimize angular deviation from the initialization. This refinement aligns the estimated kinematic parameters with the observed geometry of the part, yielding more realistic motion.

\subsection{Hidden geometry generation}
\begin{figure*}[t]
  \centering
  \setlength{\tabcolsep}{5pt}
  \begin{tabular}{@{}cccccc@{}}
        
    \textbf{Predicted 2D Mask} & \textbf{Segmented Mesh} & \textbf{Predicted Joint} & \textbf{Refined Joint} & \textbf{Interactive Object} & \textbf{Opened Object} \\
    \includegraphics[width=0.17\textwidth,height=0.15\textwidth,keepaspectratio]{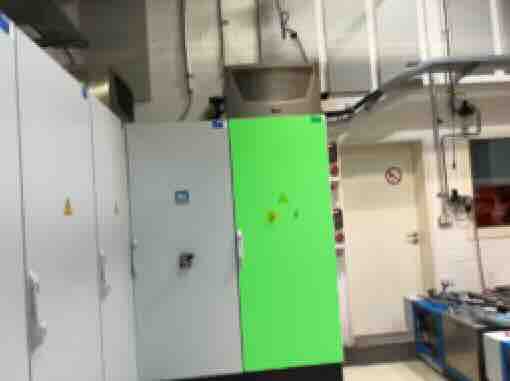} & \includegraphics[width=0.15\textwidth,height=0.15\textwidth,keepaspectratio]{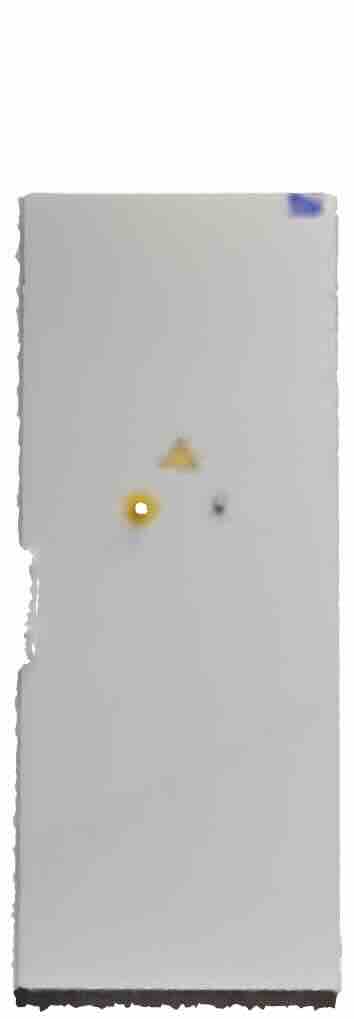} & \includegraphics[width=0.15\textwidth,height=0.15\textwidth,keepaspectratio]{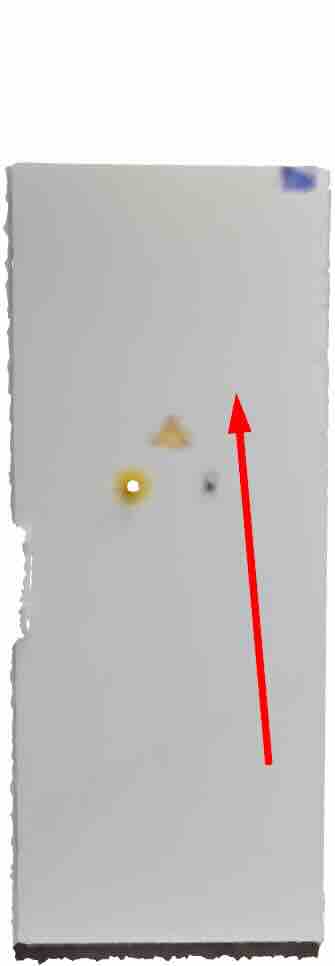} & \includegraphics[width=0.15\textwidth,height=0.15\textwidth,keepaspectratio]{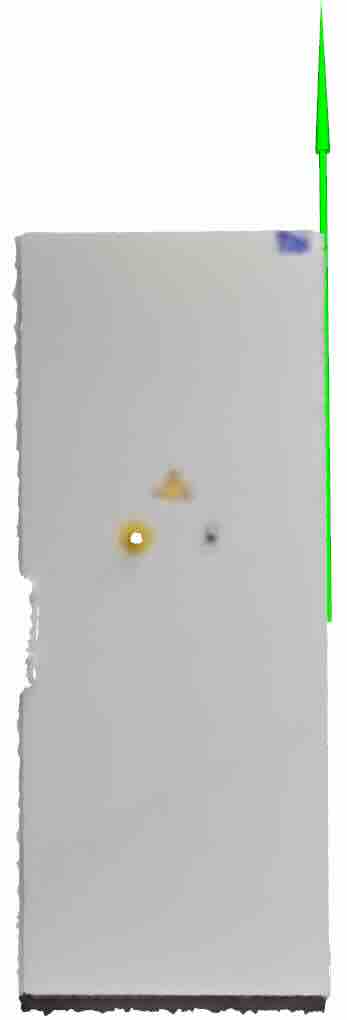} & \includegraphics[width=0.15\textwidth,height=0.15\textwidth,keepaspectratio]{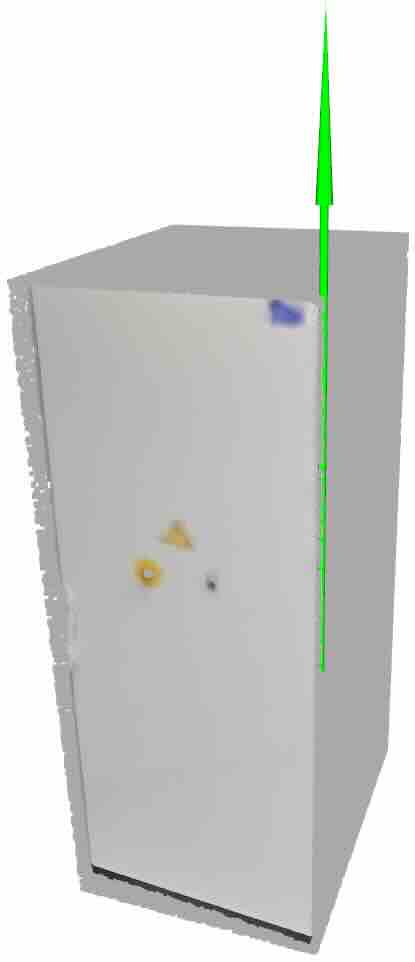} & \includegraphics[width=0.15\textwidth,height=0.15\textwidth,keepaspectratio]{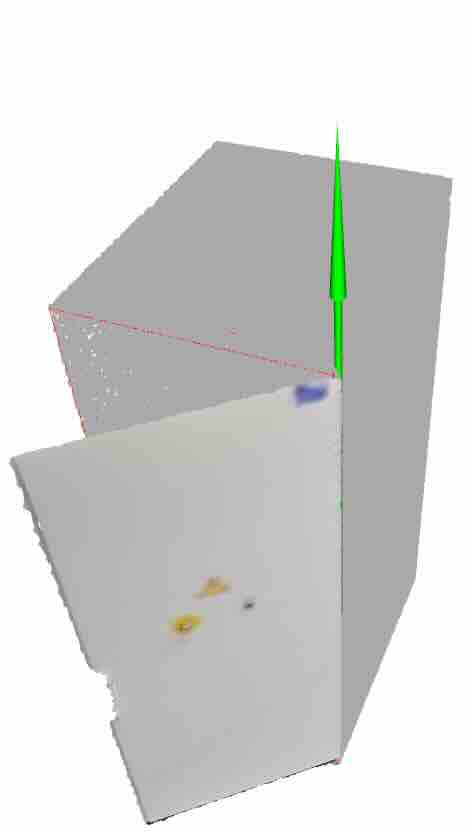}
    
  \end{tabular}
  \caption{Pipeline for interactive object generation. From left to right, the figure shows key intermediate results of interactive object generation. In the last column, the thin red line highlights the contour of the base part.}
  \label{fig:image_to_interactive_object}
  \vspace{-0.3em}
\end{figure*}

The input static 3D scene encodes only the visible surface geometry of objects. Consequently, even though articulation estimation enables object interaction, the interior behind the movable part remains unmodeled upon opening. This creates visually unrealistic artifacts and, more critically, limits practical downstream applications such as robotics, where the interior volume is essential for manipulation. To construct a complete interactive object, we generate the hidden geometry through cavity completion, producing consistent box-like structures that approximate the container behind each movable part. Fig.~\ref{fig:image_to_interactive_object} demonstrates the process of interactive object generation.

As in Sec.~\ref{subsec:articulation_estimation}, we compute an OBB for each movable part and take the face formed by its two longest edges as the front face \(P_{\mathrm{front}}\). These edges define the width and height of the inner box, while its depth is measured along the inward-pointing normal of the front face, \(\mathbf{n}_{\mathrm{front}}=\mathbf{u}_1\times \mathbf{u}_2\). We define the inner-box depth as
\[
d_{\mathrm{in}} \;=\; \min\big(\, d_{\mathrm{image}}, \; d_{\mathrm{hit}}, \; d_{\mathrm{mesh}} \,\big),
\]
and then instantiate a rectangular cavity whose front coincides with \(P_{\mathrm{front}}\) and whose extent is \(d_{\mathrm{in}}\) along \(+\mathbf{n}_{\mathrm{front}}\). Here, \(d_{\mathrm{image}}\) is an image-derived bound given by the farthest background depth in the RGB-D frame used to segment the part mesh, projected onto \(+\mathbf{n}_{\mathrm{front}}\). \(d_{\mathrm{hit}}\) is obtained by probing along \(+\mathbf{n}_{\mathrm{front}}\) from the OBB centroid to the nearest scene intersection, and taking its depth if a supporting plane can be fitted there using RANSAC. Finally, \(d_{\mathrm{mesh}}\) is the boundary depth of the static scene mesh, serving as an upper bound to prevent leakage outside the scene.

\subsection{Interactive Scene Integration}

Our goal is to produce a complete interactive scene that can be directly used in downstream applications. The output includes interactive objects consisting of movable parts with refined articulation and completed internal geometry, all without duplicates, together with the textured static background integrated into a unified representation.

\myparagraph{Duplicate removal}
To ensure clean assembly of the final interactive scene, we perform a last deduplication pass. Let \(\{S_i\}\) denote movable-part meshes with point sets \(\{X_i\}\). We compute the IoU directly on these point sets, \(\mathrm{IoU}(X_i,X_j)=\tfrac{|X_i\cap X_j|}{|X_i\cup X_j|}\). The procedure has two stages. \emph{Stage~1 (pairwise pruning):} for every pair \((i,j)\), if \(\mathrm{IoU}(X_i,X_j)\ge \tau_{\mathrm{dup}}\), we keep the mesh with greater information content (measured by point count) and discard the other, producing a reduced set \(\{\hat{S}_i\}\) with point sets \(\{\hat{X}_i\}\). \emph{Stage~2 (subdivision pruning):} duplicates can also appear when a larger mesh can be represented by the union of several smaller, complementary meshes while no single pair exceeds \(\tau_{\mathrm{dup}}\). We sort \(\{\hat{S}_i\}\) by \(|\hat{X}_i|\) in descending order and iterate from large to small so that larger parts can be explained by smaller ones first. For each \(\hat{S}_i\), we collect candidate smaller meshes  
\(\mathcal{C}_i=\{\, j : |\hat{X}_j| < |\hat{X}_i|,\ \mathrm{IoU}(\hat{X}_i,\hat{X}_j) \ge \tau_{\mathrm{low}} \,\}\), where \(\tau_{\mathrm{low}}\) is a small IoU threshold, and enumerate their combinations. If there exists a subset \(\mathcal{U}\subseteq \mathcal{C}_i\) such that \(\mathrm{IoU}\!\left(\hat{X}_i,\ \bigcup_{j\in\mathcal{U}} \hat{X}_j\right)\ge \tau_{\mathrm{dup}}\), we remove \(\hat{S}_i\) as a redundant superset. In the end, we remove all redundant movable parts together with their associated inner boxes before assembling the final interactive scene.

\myparagraph{Scene assembly}
We integrate interactive objects into the original static scene by carving the background geometry using the world-frame OBBs of deduplicated movable parts. Let the original static scene mesh be \(\mathbf{P}\) and the set of movable-part OBBs be \(\{B_j\}\). To remove residual fragments from imperfect segmentation and spurious points in the original scan, we delete all vertices lying inside any \(B_j\) together with their incident faces. This yields a cleaned background mesh
\(
\mathbf{P}_{\mathrm{rem}} \;=\; \mathbf{P} \setminus \bigcup_j B_j
\).
We then assemble the final interactive scene by combining \(\mathbf{P}_{\mathrm{rem}}\) with the interactive objects, ensuring a clean integration without interpenetration.

\myparagraph{Texture generation}
As both the original scene mesh and the segmented movable-part meshes provide only per-vertex colors, we generate texture maps based on InstantTexture~\cite{instanttexture}. We first perform UV unwrapping with Xatlas~\cite{xatlas} on the movable parts and the cleaned remainder mesh. Given the UV parameterization, we rasterize each triangle in atlas space and compute per-texel colors by barycentric interpolation of the source vertex colors, yielding dense textures aligned to the unwrapped charts. We then repair uncovered texels caused by sampling gaps or occlusions via inpainting, followed by smoothing with low-pass filtering to attenuate aliasing across chart boundaries. Finally, the generated textures are bound to their corresponding meshes and, together with the refined articulation parameters, exported as \texttt{.dae} assets that are referenced in the scene description files.

\begin{figure}[t]
  \centering
  \begin{minipage}[b]{0.02\columnwidth}
    \begin{minipage}[b]{\columnwidth}
      \centering
      \scriptsize
      \rotatebox{90}{\textbf{Static Scene}}
      \vspace{0.5em}
    \end{minipage}
    \vspace{0.5em}
    \begin{minipage}[b]{\columnwidth}
      \centering
      \scriptsize
      \rotatebox{90}{\textbf{\name{}}}
    \end{minipage}
  \end{minipage}%
  \hfill
  \begin{minipage}[b]{0.97\columnwidth}
    \begin{minipage}[b]{0.324\columnwidth}
      \includegraphics[width=\linewidth,height=0.6\linewidth,keepaspectratio]{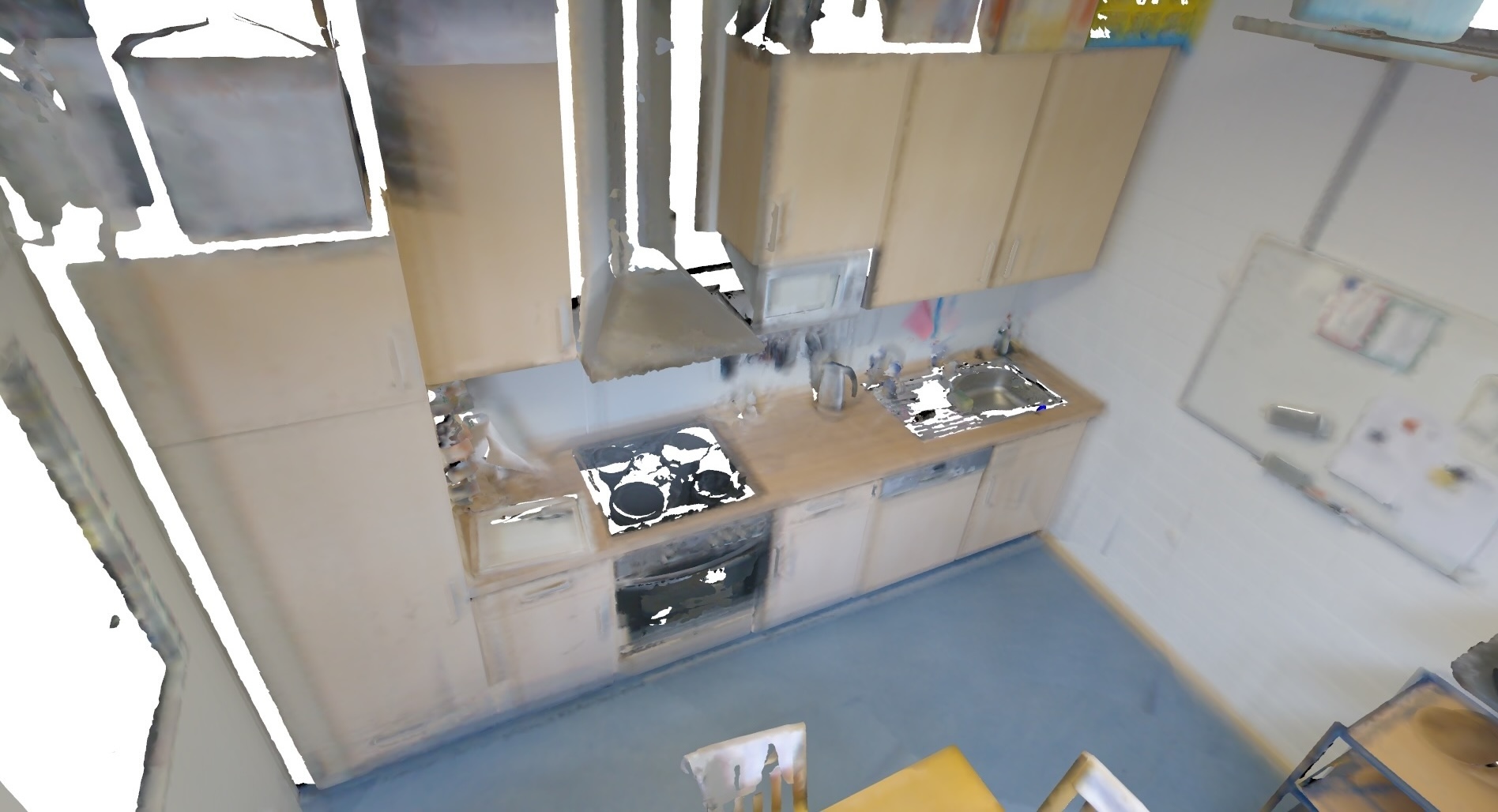}
    \end{minipage}
    \hfill
    \begin{minipage}[b]{0.324\columnwidth}
      \includegraphics[width=\linewidth,height=0.6\linewidth,keepaspectratio]{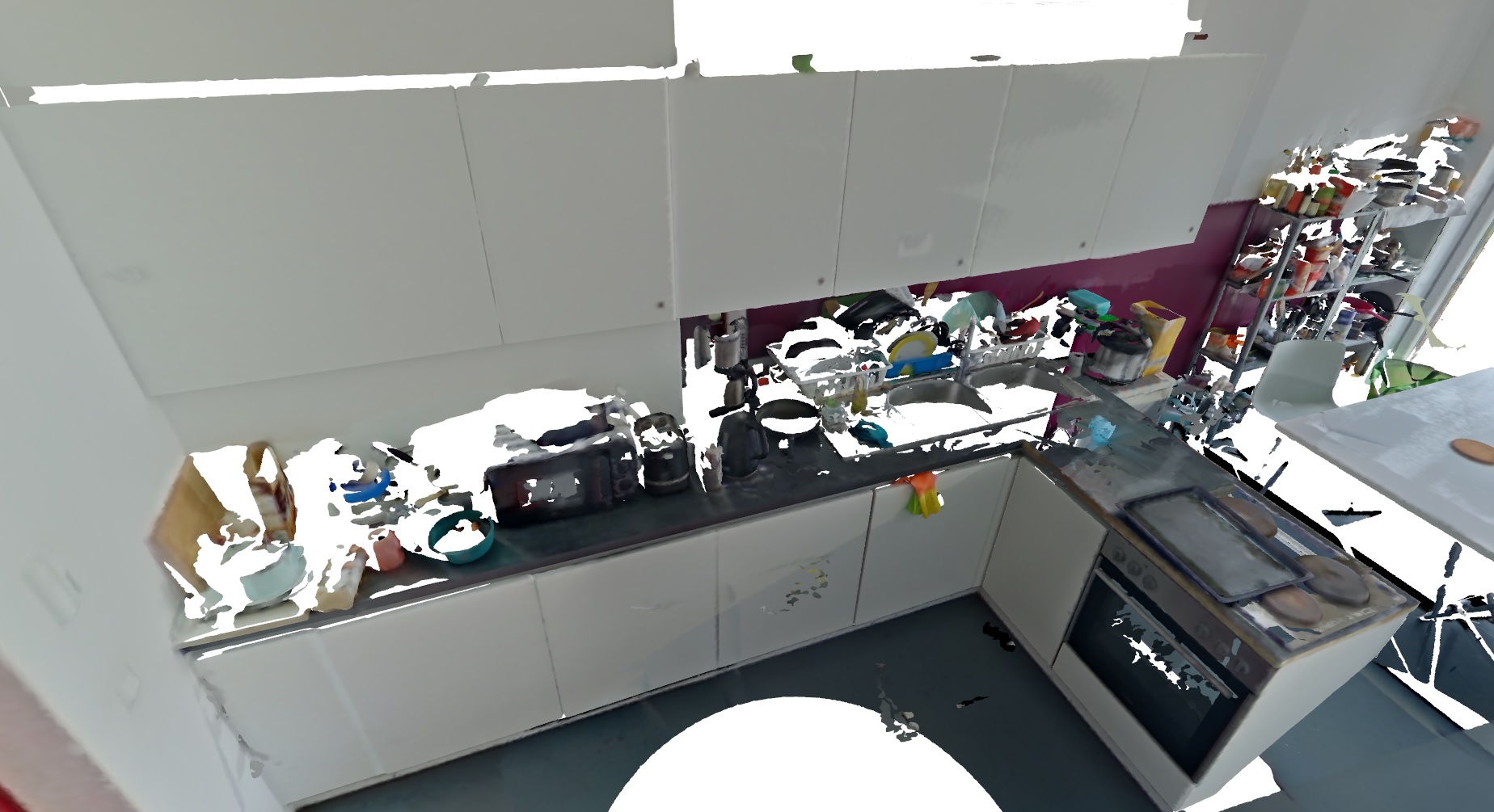}
    \end{minipage}
    \hfill
    \begin{minipage}[b]{0.324\columnwidth}
      \includegraphics[width=\linewidth,keepaspectratio]{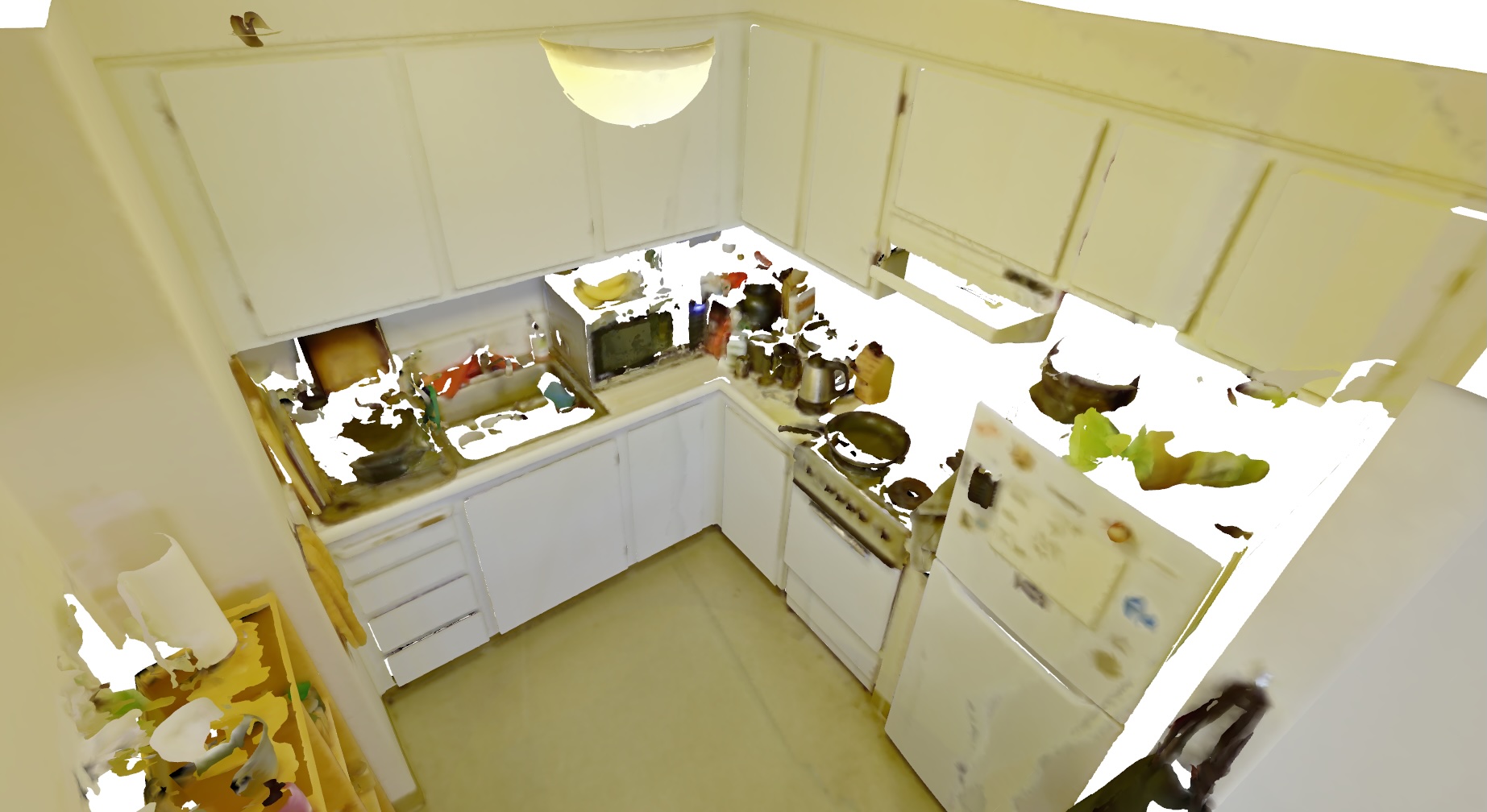}
    \end{minipage}

    \vspace{0.35em}

    \begin{minipage}[b]{0.324\columnwidth}
      \includegraphics[width=\linewidth,height=0.6\linewidth,keepaspectratio]{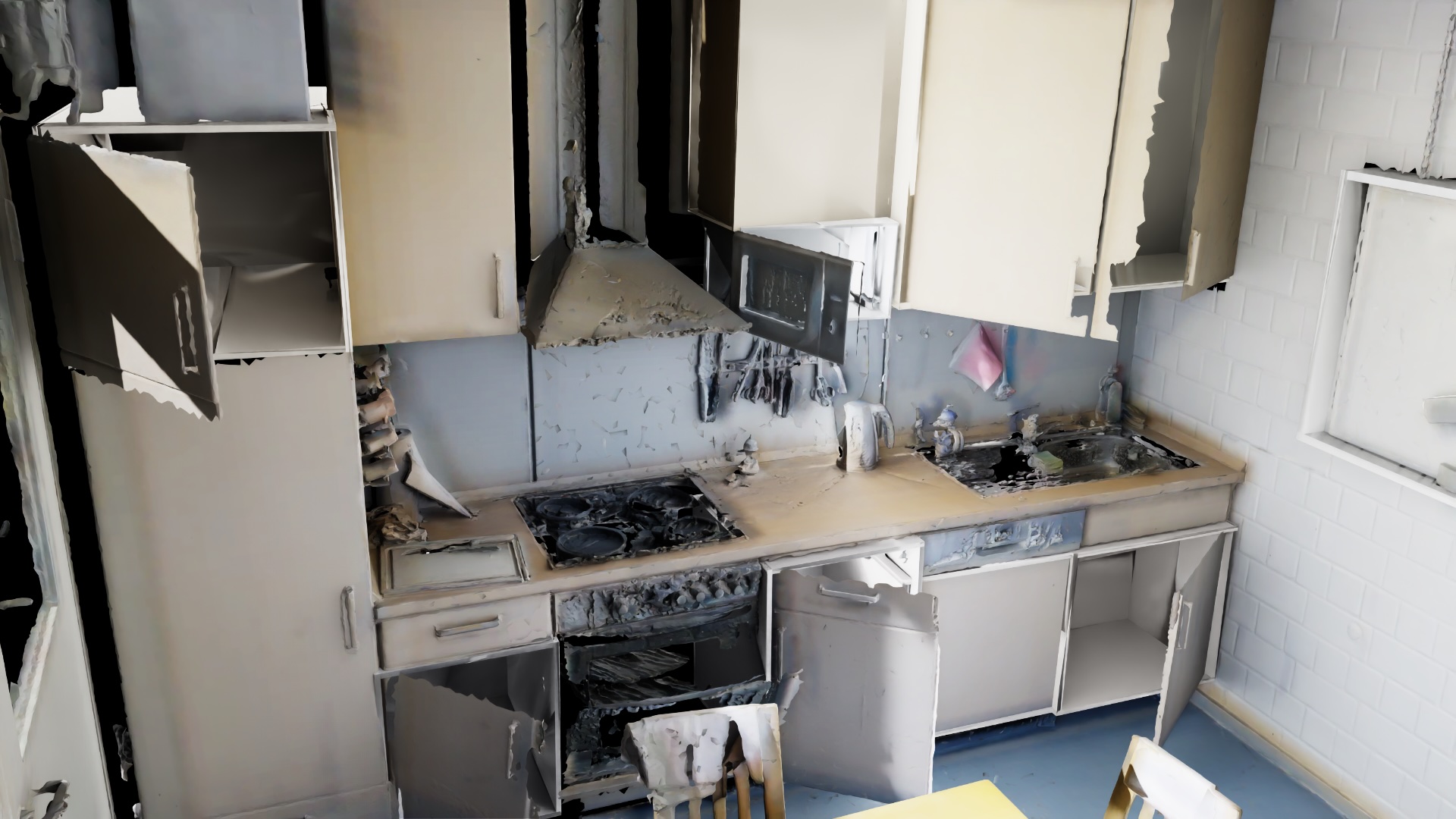}
    \end{minipage}
    \hfill
    \begin{minipage}[b]{0.324\columnwidth}
      \includegraphics[width=\linewidth,keepaspectratio]{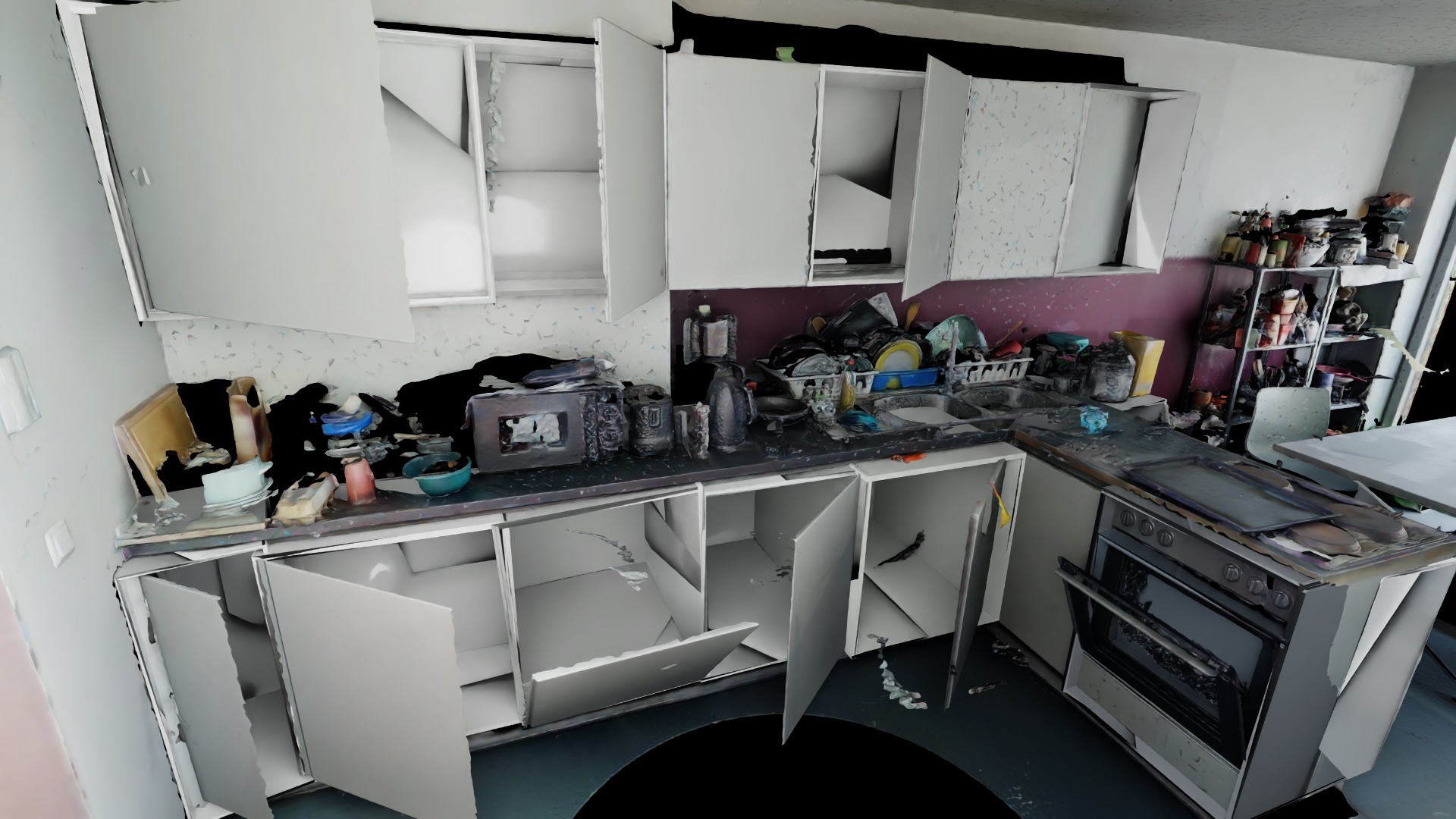}
    \end{minipage}
    \hfill
    \begin{minipage}[b]{0.324\columnwidth}
      \includegraphics[width=\linewidth,keepaspectratio]{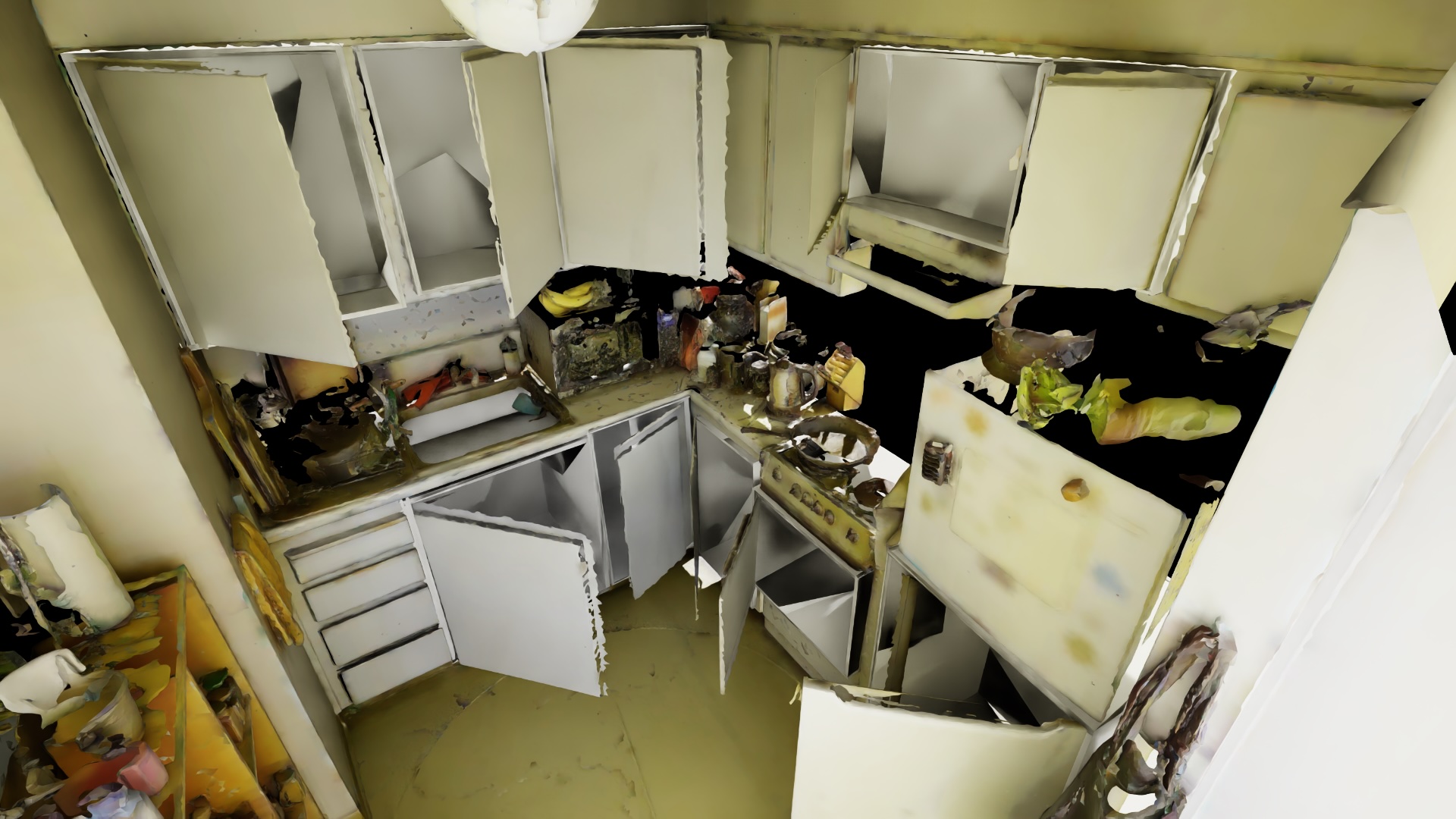}
    \end{minipage}
  \end{minipage}

  \caption{Qualitative results of \name{}.
  Static input scenes from ScanNet++ and the interactive outputs generated by \name{}, visualized in Isaac Sim.}
  \label{fig:isaacsim_result_ours}
\end{figure}

\begin{figure}[t]
  \centering
  \begin{minipage}[b]{0.99\columnwidth}
    \begin{minipage}[b]{0.495\columnwidth}
      \centering
      \footnotesize\textbf{ROS}\\[0.3em]
      \includegraphics[width=\linewidth,height=0.55\linewidth]{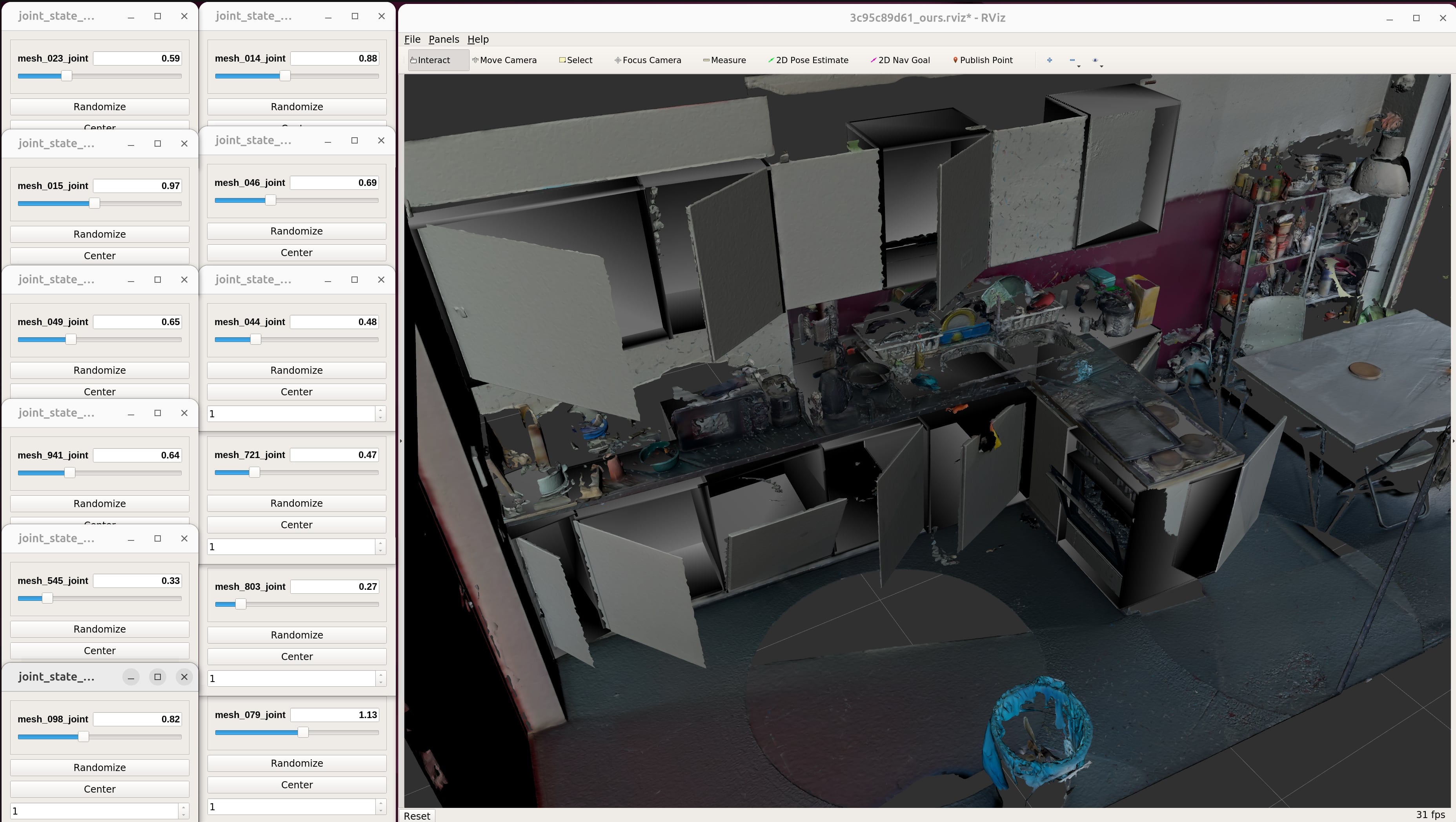}
    \end{minipage}
    \hfill
    \begin{minipage}[b]{0.495\columnwidth}
      \centering
      \footnotesize\textbf{Isaac Sim}\\[0.3em]
      \includegraphics[width=\linewidth,height=0.55\linewidth]{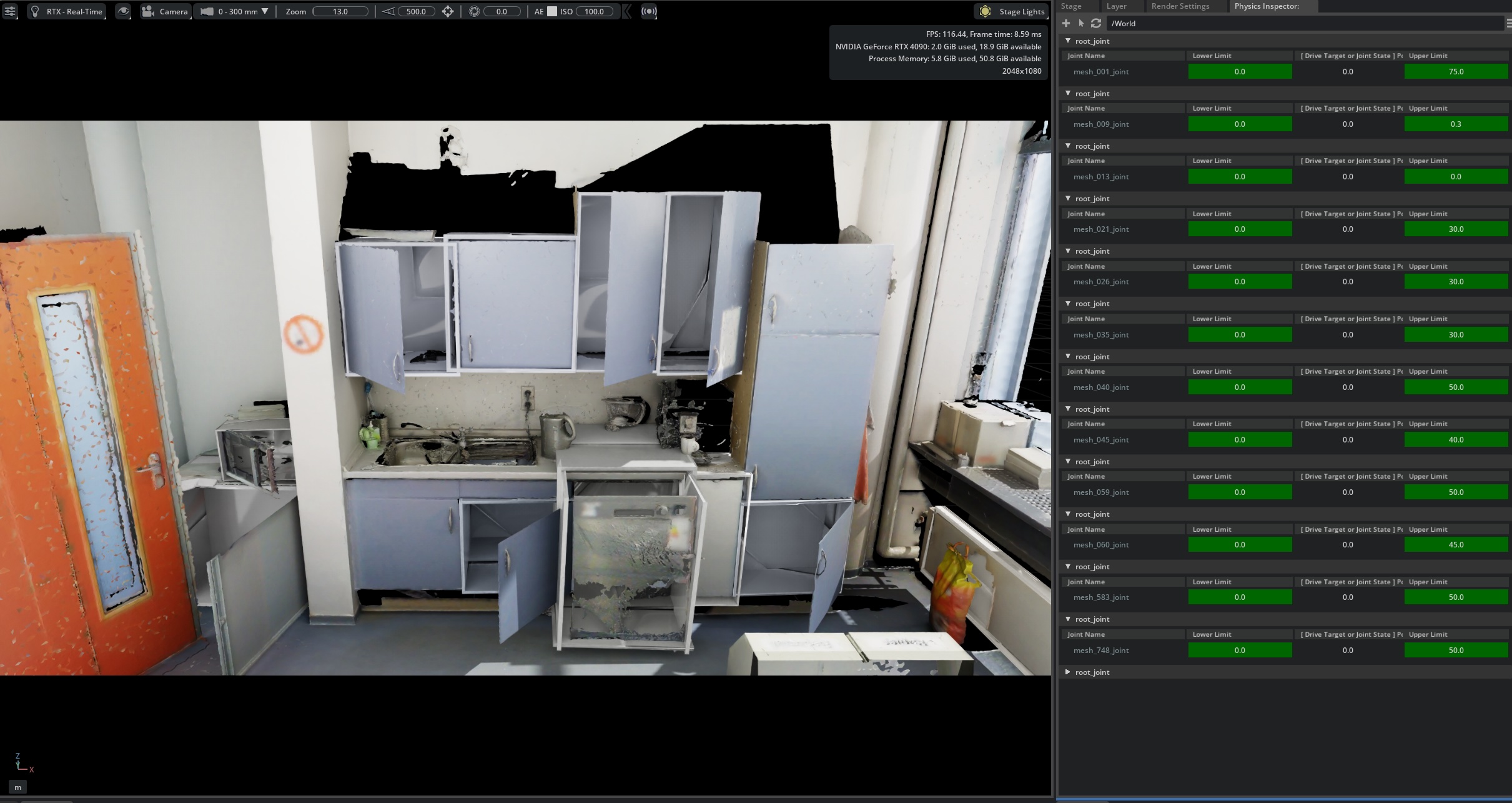}
    \end{minipage}
  \end{minipage}

    
  \caption{Manipulation GUIs. Interfaces in ROS and Isaac Sim enabling per-object articulation control and benchmarking.}
  \label{fig:gui}
\end{figure}

\myparagraph{Export and simulator support}
To ensure portability across visualization and simulation stacks, we export each interactive scene in two widely adopted formats: \emph{URDF} for the robotics ecosystem and \emph{USD} for graphics/physics engines. The exports package the cleaned background, articulated interactive objects, and textures, enabling direct deployment in PyBullet~\cite{coumans2021}, ROS~\cite{quigley2009ros}, and Isaac~Sim~\cite{NVIDIA_Isaac_Sim}. Fig.~\ref{fig:isaacsim_result_ours} shows the simulated interactive scenes generated by \name{} from different inputs in Isaac Sim. For ROS, we additionally provide a lightweight GUI for manipulating the interactive scene during visualization and benchmarking (see Fig.~\ref{fig:gui}). 

\begin{table*}[!t]
  \caption{Quantitative results on openable object detection. Openable object detection on 30 ScanNet++ scenes rich in openable objects. We report Precision, Recall, and F1 at IoU thresholds $\tau\!\in\!\{0.25,0.50\}$. 
  ${}^{*}$For URDFormer, $\tau$ denotes the coverage ratio, while for other methods it denotes IoU. $^{\dagger}$For URDFormer, the number reflects only openable objects visible in the selected region with the highest density of openable objects, since the method takes a single RGB image as input.}

  \centering
  \resizebox{\linewidth}{!}{
  \begin{tabular}{
    @{} l c 
    S[table-format=1.3] S[table-format=1.3] S[table-format=1.3]
    S[table-format=1.3] S[table-format=1.3] S[table-format=1.3]
    @{} }
    \toprule
    \multirow{3}{*}{\textbf{Method}} & \multirow{3}{*}{\textbf{\#Openable Objects}} &
      \multicolumn{3}{c}{\boldmath$\tau^{*}=0.25$} &
      \multicolumn{3}{c}{\boldmath$\tau^{*}=0.50$} \\
    \cmidrule(lr){3-5} \cmidrule(lr){6-8} 
    & & \textbf{Precision\,$\uparrow$} & \textbf{Recall\,$\uparrow$} & \textbf{F1 Score\,$\uparrow$}
      & \textbf{Precision\,$\uparrow$} & \textbf{Recall\,$\uparrow$} & \textbf{F1 Score\,$\uparrow$} \\
    \midrule

    URDFormer   & 226$^{\dagger}$  & 0.429 & 0.226 & 0.296 & 0.361 & 0.190 & 0.249 \\
    DRAWER (GT-mesh)   & 540  & 0.434 & 0.141 & 0.213 & 0.326 & 0.106 & 0.159 \\
    \name{}    & 540  & \best{0.731} & \best{0.393} & \best{0.511} & \best{0.666} & \best{0.357} & \best{0.465} \\
    

    
    \bottomrule
  \end{tabular}}
    \par\vspace{3pt}
    \makebox[\linewidth][l]{\footnotesize\hspace{0.3em} The best result for each metric is highlighted in \textbf{bold}.}
  \label{tab:eval_detection}
\end{table*}

\section{Experiments}
\label{sec:experiments}

\subsection{Experimental Setup}
\label{subsec:experimental_setup}
\myparagraph{Datasets}
To comprehensively evaluate our method and baselines, we use ScanNet++~\cite{yeshwanthliu2023scannetpp}, a widely adopted dataset that provides high-quality posed RGB-D reconstructions across diverse indoor environments. We select 30 scenes rich in articulated objects, ranging from single rooms to full apartments. For quantitative evaluation, we adopt Articulate3D~\cite{halacheva2024articulate3d} as ground truth, which augments ScanNet++ with part-level semantic labels and articulation annotations (joint type, origin, axis, and motion range). Our evaluation focuses on objects with open-close motion, while small controls such as buttons and knobs are excluded. For benchmarking, we evaluate our method against the baselines URDFormer and DRAWER.

\myparagraph{Evaluation metrics}
We assess performance along two dimensions: movable-part detection and articulation parameters. For detection, we compute vertex-level IoU over all pairs of predicted movable-part meshes and ground-truth part meshes from Articulate3D in world coordinates. As the predicted parts are segmented directly from the input static scene, both predictions and ground truth reside in the same coordinate frame, requiring no additional alignment. A prediction is counted as a true positive if its IoU with at least one ground-truth part exceeds \(\tau\). We report precision, recall, and F1 at \(\tau\in\{0.25,0.50\}\) across all scene objects.

For articulation, we consider only true positive detections, evaluating each predicted joint against its matched ground-truth counterpart. We first assess the predicted joint type and report joint-type accuracy. Following the MultiScan protocol~\cite{mao2022multiscan}, articulation is evaluated for parts with a correct type match using two metrics: (i) \emph{Minimum Distance (MD)}, the shortest distance between the predicted and ground-truth joint lines, which better captures positional error than origin-to-origin distance since origin differences along the joint axis do not alter the induced motion; and (ii) \emph{Orientation Error (OE)}, the angle between the predicted and ground-truth joint axes, capturing directional error. As prismatic motion is origin-invariant, we report only OE for prismatic joints, while both MD and OE are reported for revolute joints.

To ensure fair comparison across methods with different articulation priors, we adopt a three-way joint-type taxonomy: prismatic, horizontal revolute, and vertical revolute. Most indoor revolute mechanisms are near-vertical, so methods that always predict vertical axes gain an artificial advantage if the revolute joint is treated as a single class in evaluation, yielding deceptively low OE. Conversely, methods that explicitly model horizontal axes can be penalized with errors near \SI{90}{\degree}. To mitigate this bias, we classify a revolute joint as horizontal if its axis forms less than \SI{45}{\degree} with the ground plane, and vertical otherwise. Joint-type accuracy and OE are reported under this taxonomy, revealing true motion behavior while isolating differences in method capability.

\begin{figure*}[t]
  \centering
  \begin{minipage}[b]{\textwidth}
      \begin{minipage}[b]{0.245\textwidth}
        \centering
        \textbf{Input$^{\ast}$}\\[0.3em]
        \includegraphics[width=\linewidth,height=0.6\linewidth,keepaspectratio]{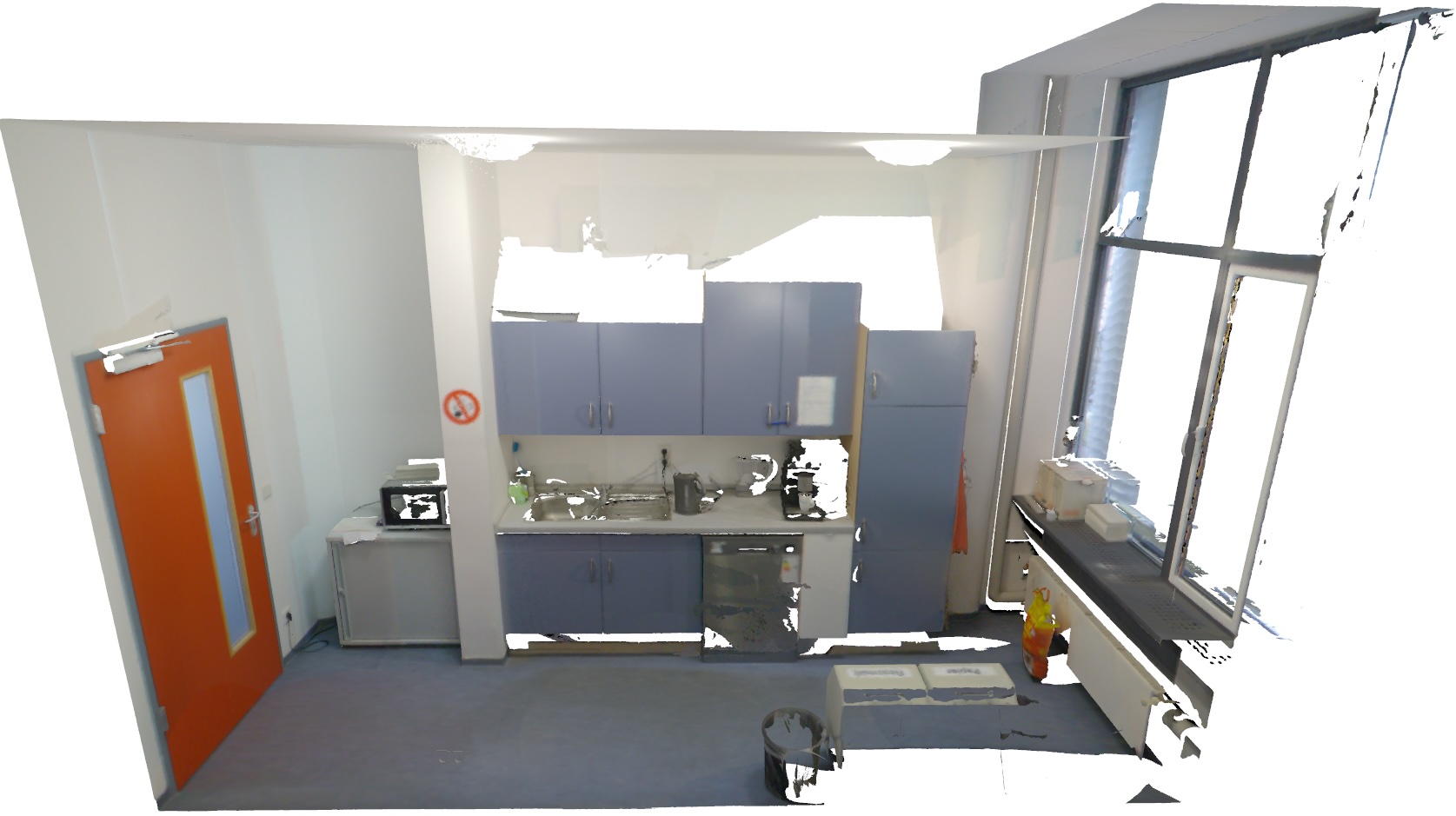}
      \end{minipage}
      \hfill
      \begin{minipage}[b]{0.245\textwidth}
        \centering
        \textbf{URDFormer}\\[0.3em]
        \includegraphics[width=\linewidth,height=0.56\linewidth,keepaspectratio]{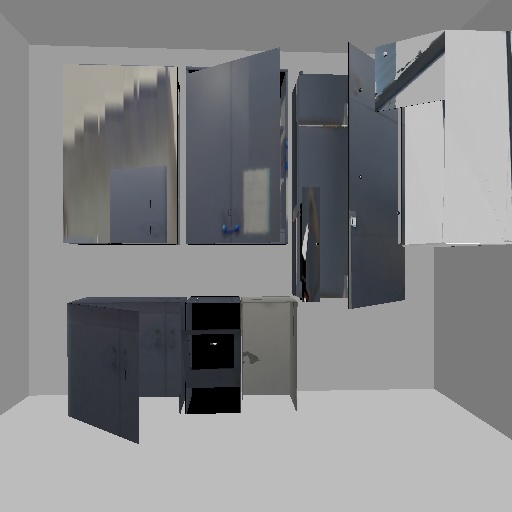}
      \end{minipage}
      \hfill
      \begin{minipage}[b]{0.245\textwidth}
        \centering
        \textbf{DRAWER}\\[0.3em]
        \includegraphics[width=\linewidth,keepaspectratio]{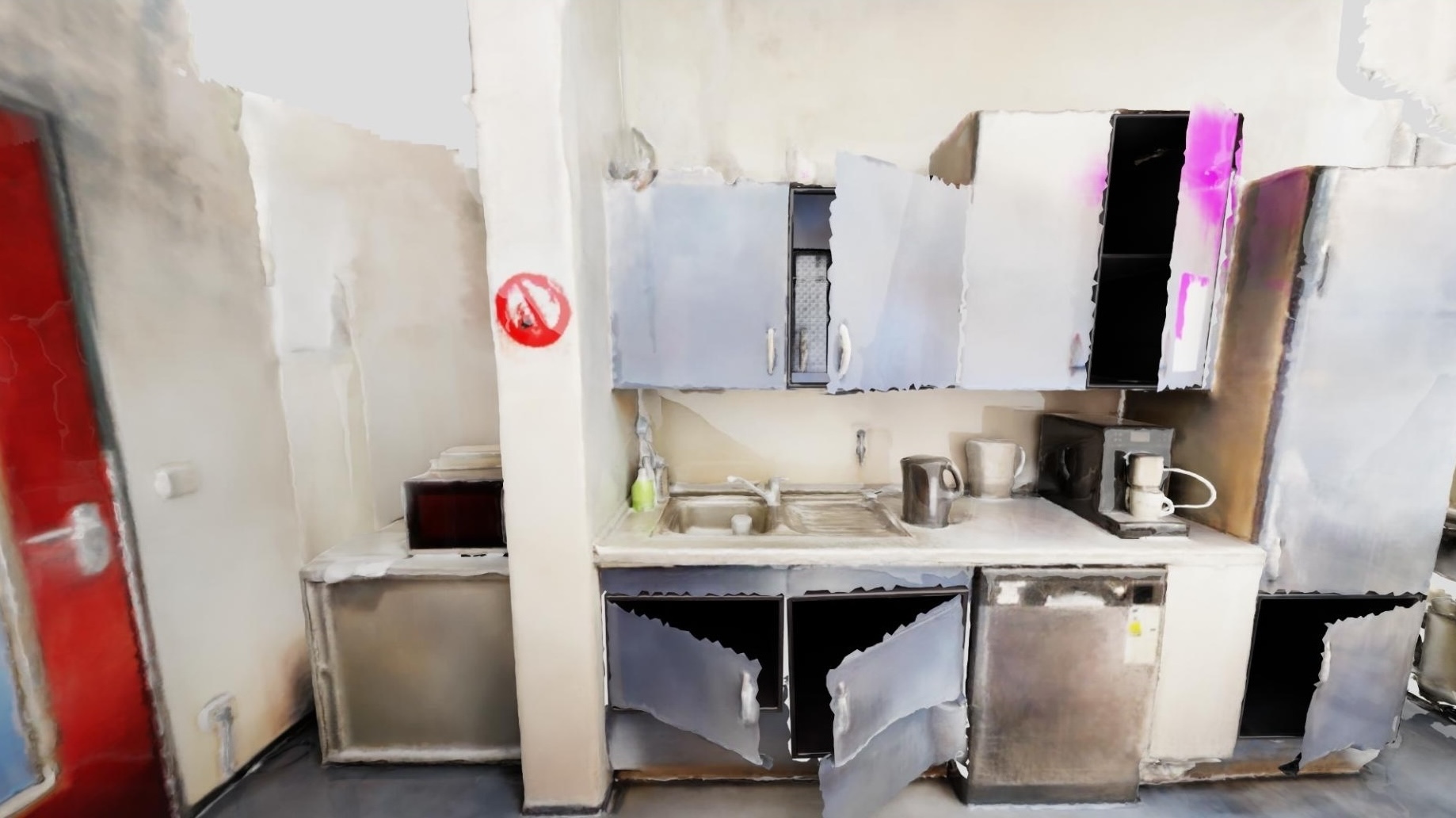}
      \end{minipage}
      \hfill
      \begin{minipage}[b]{0.245\textwidth}
        \centering
        \textbf{\name{}}\\[0.3em]
        \includegraphics[width=\linewidth,keepaspectratio]{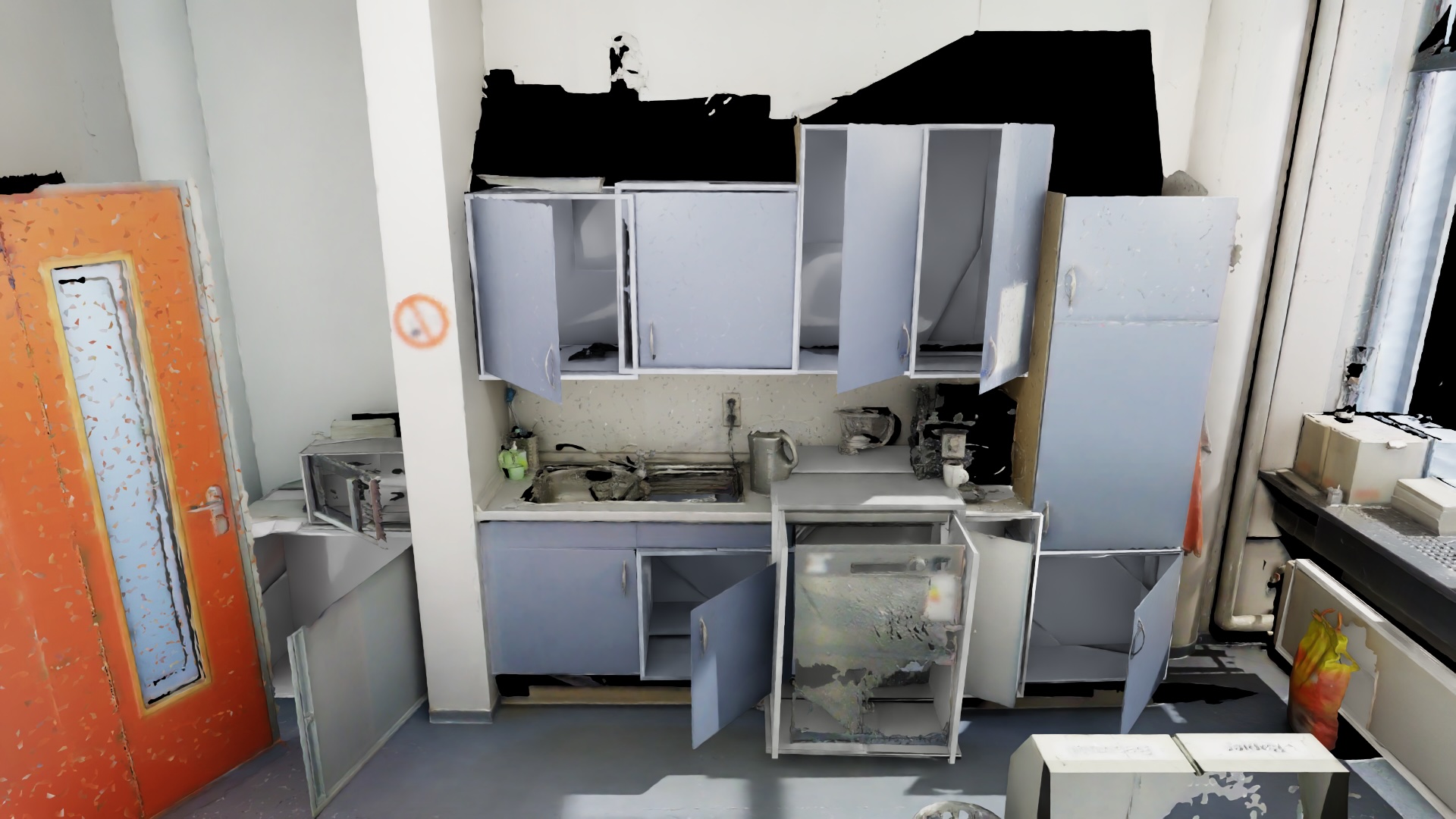}
      \end{minipage}
  \end{minipage}
    \caption{Qualitative comparison of generated interactive scenes. For each input static scene, we show results produced by each native method without any adaptation or modification. Visualizations follow each method's native support: URDFormer is rendered in PyBullet, while DRAWER and \name{} are rendered in Isaac Sim. $^{\ast}$Input corresponds to the ScanNet++ scene mesh, the posed RGB-D frames, or a captured image of the scene.}
  \label{fig:final_results_comparison}
\end{figure*}

\begin{table*}[!t]
  \caption{Quantitative results on articulation estimation and MOD evaluation. Articulation evaluation on the same 30 scenes. Joint type accuracy is computed over correctly detected movable parts, while minimum distance (MD) and orientation error (OE) are reported only for parts with correctly predicted joint types. ${}^{*}$${}^{\dagger}$ Conditions are the same as in Tab.~\ref{tab:eval_detection}.}

  \centering
  \resizebox{\linewidth}{!}{
  \begin{tabular}{
      @{} l c
      S[table-format=2.1]  
      S[table-format=1.3]  
      S[table-format=1.2]  
      S[table-format=2.1] S[table-format=2.1] S[table-format=2.1] S[table-format=2.1]
      S[table-format=2.1]  
      S[table-format=1.3]  
      S[table-format=1.2]  
      S[table-format=2.1] S[table-format=2.1] S[table-format=2.1] S[table-format=2.1]
      @{}}
    \toprule
    \multirow{3}{*}{\textbf{Dataset/Method}} & \multirow{3}{*}{\makecell{\#\textbf{Openable}\\\textbf{Objects}}} &
      \multicolumn{7}{c}{\boldmath$\tau^{*}=0.25$} &
      \multicolumn{7}{c}{\boldmath$\tau^{*}=0.50$} \\
    \cmidrule(lr){3-9} \cmidrule(lr){10-16}
    & &
      \multicolumn{3}{c}{\textbf{Articulation}} &
      \multicolumn{4}{c}{\textbf{MOD\,[\si{\percent}]$\uparrow$}} &
      \multicolumn{3}{c}{\textbf{Articulation}} &
      \multicolumn{4}{c}{\textbf{MOD\,[\si{\percent}]$\uparrow$}} \\
    \cmidrule(lr){3-5} \cmidrule(lr){6-9} \cmidrule(lr){10-12} \cmidrule(lr){13-16}
    & & \textbf{Joint Acc.\,[\si{\percent}]$\uparrow$} & \textbf{MD\,[\si{m}]$\downarrow$} & \textbf{OE\,[\si{\degree}]$\downarrow$}
      & \textbf{PDet} & \textbf{+M} & \textbf{+MO} & \textbf{+MOD}
      & \textbf{Joint Acc.\,[\si{\percent}]$\uparrow$} & \textbf{MD\,[\si{m}]$\downarrow$} & \textbf{OE\,[\si{\degree}]$\downarrow$}
      & \textbf{PDet} & \textbf{+M} & \textbf{+MO} & \textbf{+MOD} \\
    \midrule

    URDFormer        & 226$^{\dagger}$ & 51.0 & 0.583 & \best{0.50} & 22.6 & 11.5 & 11.5 & 2.2 & 46.5 & 0.491 & \best{0.61} & 19.0 & 8.8 & 8.8 & 3.1 \\
    DRAWER (GT-mesh) & 540 & 73.7 & 0.288 & 1.62 & 14.1 & 10.4 & 10.0 & 5.2 & 78.9 & 0.313 & 1.71 & 10.6 & 8.3 & 8.0 & 3.7 \\
    \name{}             & 540 & \best{91.5} & \best{0.203} & 1.16 & \best{39.3} & \best{35.9} & \best{35.9} & \best{25.2} & \best{92.2} & \best{0.180} & 1.13 & \best{35.7} & \best{33.0} & \best{33.0} & \best{23.9} \\
    


    
    \bottomrule
  \end{tabular}}
  \par\vspace{3pt}
    \makebox[\linewidth][l]{\footnotesize\hspace{0.2em} The best result for each metric is highlighted in \textbf{bold}.}
  \label{tab:eval_combined}
\end{table*}

\begin{figure*}[tbp]
  \centering

  \begin{minipage}[b]{0.24\linewidth}
    \centering
    \includegraphics[width=\linewidth]{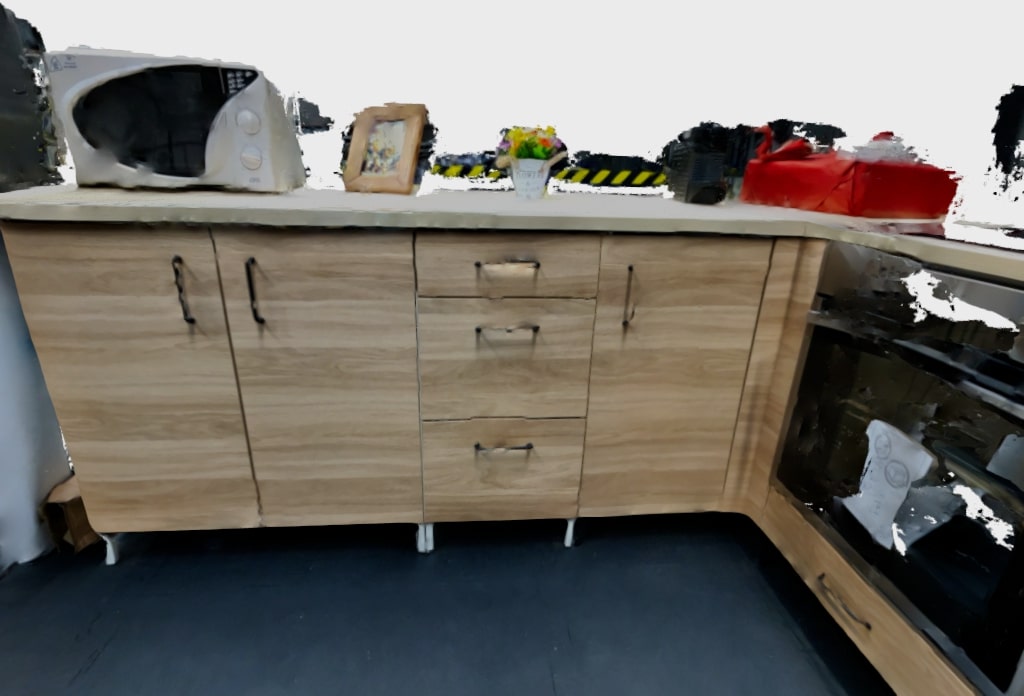}
    \\
    {\scriptsize (a) Static Scan}
  \end{minipage}%
  \hfill
  \begin{minipage}[b]{0.24\linewidth}
    \centering
    \includegraphics[width=\linewidth]{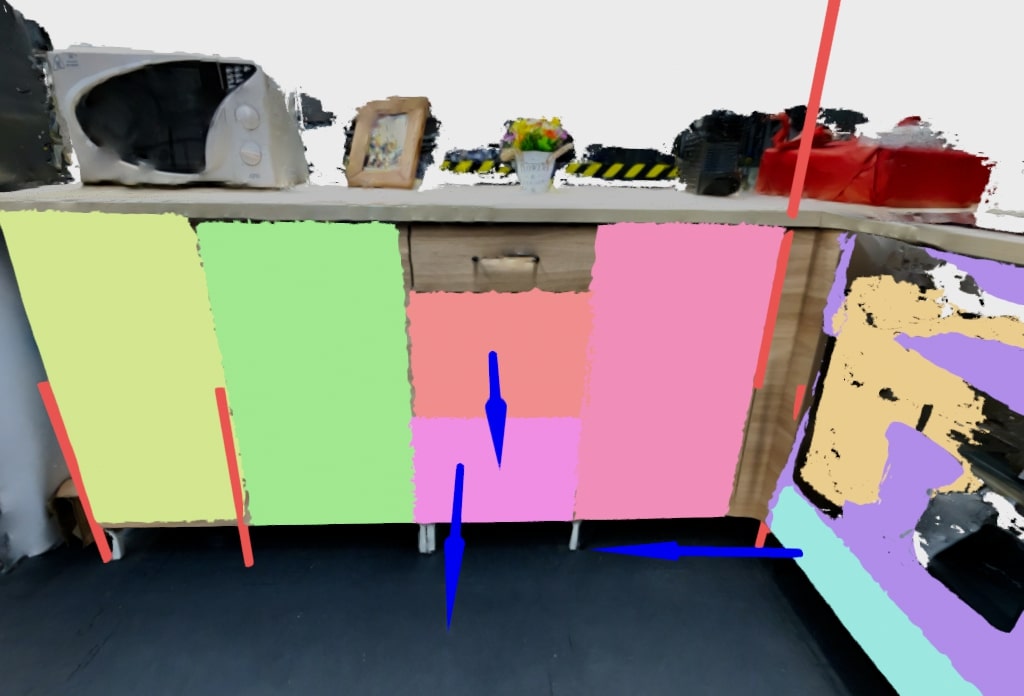}
    \\
    {\scriptsize (b) Output}
  \end{minipage}%
  \hfill
  \begin{minipage}[b]{0.24\linewidth}
    \centering
    \includegraphics[width=\linewidth]{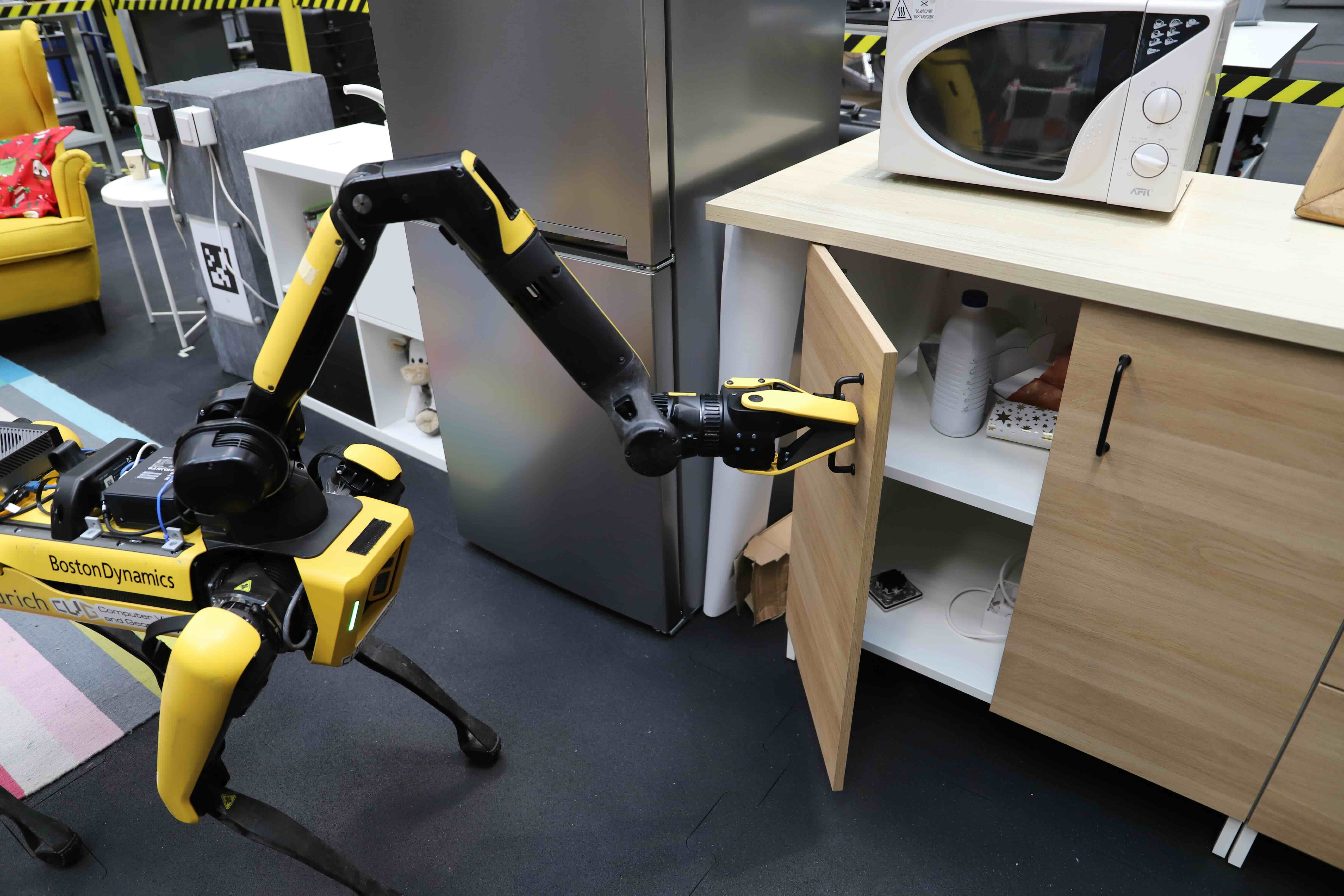}
    \\
    {\scriptsize (c) Revolute}
  \end{minipage}%
  \hfill
  \begin{minipage}[b]{0.24\linewidth}
    \centering
    \includegraphics[width=\linewidth]{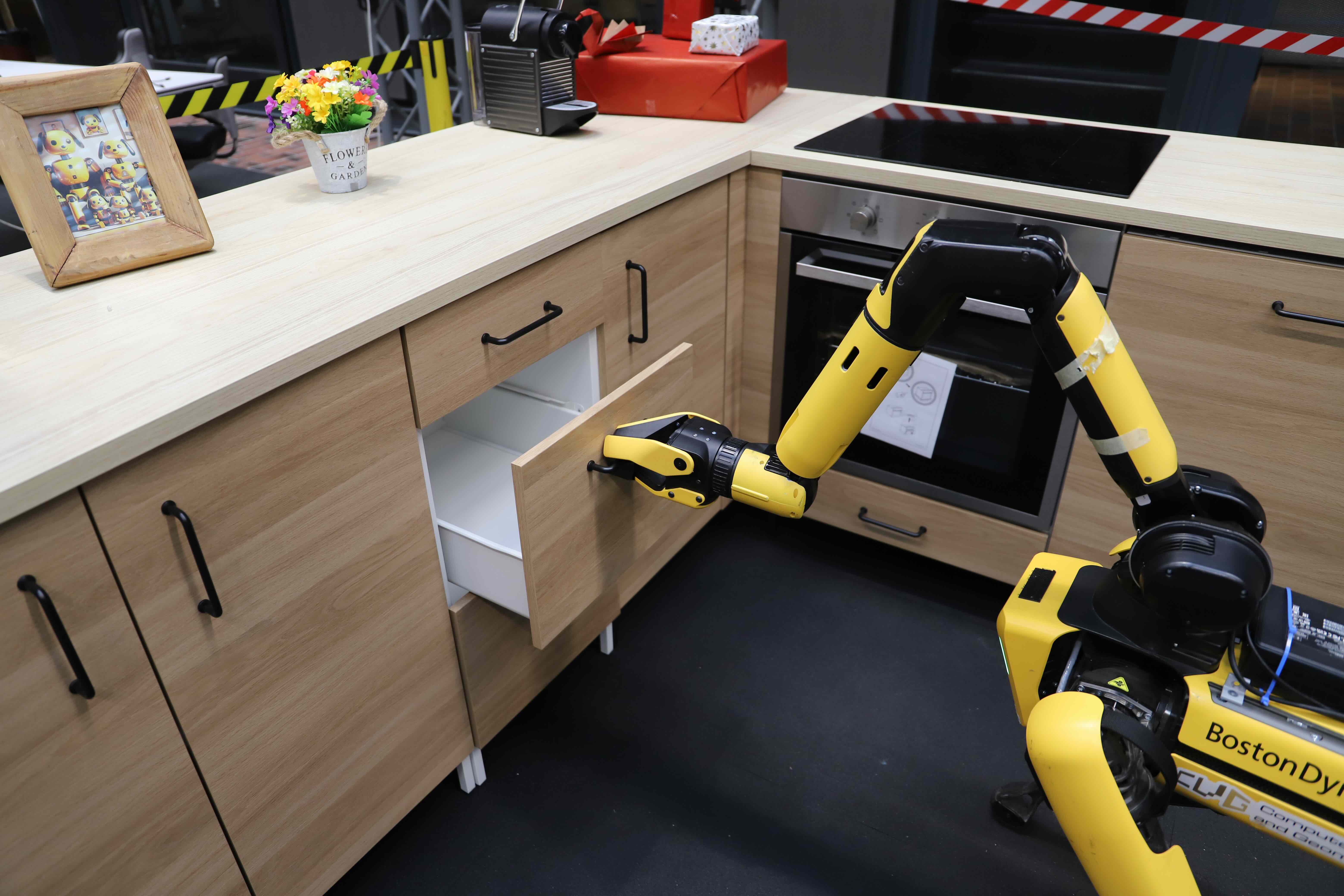}
    \\
    {\scriptsize (d) Prismatic}
  \end{minipage}

    
  \caption{Real-world Experiment. Interactive scene generated from a real-world scan, enabling robotic manipulation of revolute and prismatic joints. }

  \label{fig:real2sim2real}
\end{figure*}

\myparagraph{Implementation Details}
The threshold values utilized in our pipeline were determined empirically and stay static across all experiments. We conducted a sensitivity analysis on the parameters of the non-learning algorithms, which demonstrates the robustness of our parameter choices.

\subsection{Baselines}
\label{subsec:evaluation_method}
\noindent\emph{URDFormer}~\cite{chen2024urdformerpipelineconstructingarticulated} generates an interactive scene from a single RGB image without guaranteed metric scale or an explicit mapping to the source 3D scene. For fair comparison, we adopt a concise placement-scaling protocol. For each static scene, we select a region rich in potentially interactive objects and capture a posed RGB-D image. We run the official model, extract per-object 3D bounding boxes and a global box, estimate a global scale \(s\) as the ratio between the vertical extent of the ground-truth region and that of the predicted global box, then scale and place the prediction back into the static scene using the pose and depth. As the output is not segmented from the scene mesh, vertex-level IoU is inapplicable. We therefore evaluate detection via 3D coverage: for a predicted box \(B\) and a ground-truth movable-part mesh \(M\) with vertices \(V(M)\) in world coordinates, \(\mathrm{cov}(B,M)=\frac{|V(M)\cap B|}{|V(M)|}\) and a part is detected if \(\max_B \mathrm{cov}(B,M)\ge \tau\). Articulation is then evaluated with the same protocol and metrics defined above.

\noindent\emph{DRAWER}~\cite{xia2025drawerdigitalreconstructionarticulation} reconstructs a scene mesh from RGB frames via monocular depth and normal estimation and generates an interactive scene on top. For fair comparison on ScanNet++, we bypass reconstruction and directly use ground-truth meshes as input. All downstream modules follow official settings and pretrained weights, except that the original GPT-4o component is replaced with Gemini~2.5~Flash. We denote this configuration as \emph{DRAWER (GT-mesh)}. This protocol aligns the input domain with ours, ensuring independence from reconstruction errors in evaluation.

\subsection{Results}
\myparagraph{Openable object detection}
We first assess how many openable objects each method recovers in the generated interactive scenes. Table~\ref{tab:eval_detection} compares URDFormer, DRAWER (GT-mesh), and \name{} under the experimental setup. Note that the output scenes of URDFormer are not at true scale as explained in Sec.~\ref{subsec:evaluation_method}, whereas the other methods produce scenes at true scale. Across all metrics, our framework achieves the best results and substantially outperforms both baselines, demonstrating its superior ability to consistently and accurately recover interactive objects across diverse indoor scenes.

\addtolength{\textheight}{-0.4cm}

\myparagraph{Articulation estimation}
As shown in Tab.~\ref{tab:eval_combined}, our approach achieves over 90\% joint-type accuracy, substantially surpassing all baselines. It also attains the lowest MD and OE among methods with learned articulation, outperforming DRAWER, the prior state of the art with true-scale results. Although URDFormer reports the lowest OE, this is due to its heuristic strategy of presetting articulation axes to vertical or horizontal relative to the ground based on joint type. Fig.~\ref{fig:part_arti_comparison} shows scene-level comparisons of movable-part detection and articulation results, while Fig.~\ref{fig:final_results_comparison} compares interactive scenes generated by native methods.

We further report results under the MOD setting in Tab.~\ref{tab:eval_combined}, following the same experimental setup. This setting provides a cumulative view of articulation parameter correctness and reveals which metric is the primary performance bottleneck. Here, \textbf{PDet} measures detection accuracy, while \textbf{+M}, \textbf{+MO}, and \textbf{+MOD} denote progressively stricter criteria: correct motion type, motion type with OE $< \SI{10}{\degree}$, and additionally MD $< 0.25 \times$ the part mesh diagonal, respectively~\cite{mao2022multiscan}. Each stage is evaluated only if all preceding conditions are satisfied.

\myparagraph{Ablation study}
To assess the contribution of our articulation refinement module, we conduct an ablation study by comparing results with and without it. As shown in Tab.~\ref{tab:eval_ablation}, refinement consistently improves performance across both IoU thresholds. The reduction in minimum distance demonstrates more accurate joint localization, while the dramatic decrease in orientation error underscores its importance for reliable axis direction estimation. Overall, the results demonstrate that refinement is crucial for producing precise articulation parameters and achieving robust interactive scene generation.

    
    
    
    

\begin{table}
    \caption{Ablation study on articulation refinement. minimum distance (MD) and orientation error (OE) are reported at $\tau\!\in\!\{0.25,0.50\}$. Refinement consistently enhances accuracy, with notably large improvements in OE.}
  \centering
  \resizebox{\linewidth}{!}{
  \begin{tabular}{
    @{} l
    S[table-format=1.3] S[table-format=1.2]
    S[table-format=1.3] S[table-format=1.2]
    @{}}
    \toprule
    \multirow{2}{*}{\textbf{Method}} &
      \multicolumn{2}{c}{\boldmath$\tau=0.25$} &
      \multicolumn{2}{c}{\boldmath$\tau=0.50$} \\
    \cmidrule(lr){2-3} \cmidrule(lr){4-5}
    & \textbf{Avg.\,MD\,[m]$\downarrow$} & \textbf{Avg.\,OE\,[\si{\degree}]$\downarrow$}
    & \textbf{Avg.\,MD\,[m]$\downarrow$} & \textbf{Avg.\,OE\,[\si{\degree}]$\downarrow$} \\
    \midrule
    w/o refinement & 0.226 & 6.72 & 0.213 & 6.64 \\
    w/ refinement  & \best{0.203} & \best{1.16} & \best{0.180} & \best{1.13} \\
    \bottomrule
  \end{tabular}}
  \par\vspace{3pt}
    \makebox[\linewidth][l]{\footnotesize\hspace{0.3em} The best result for each metric is highlighted in \textbf{bold}.}
  \label{tab:eval_ablation}
\end{table}

\myparagraph{Runtime analysis} 
Our framework demonstrates significantly higher computational efficiency compared to the official DRAWER baseline. In a representative scene, our approach requires 33,500s to process the full RGB-D sequence, whereas DRAWER takes 43,638s, which is $\sim$30\% longer. Notably, our current sequential implementation offers substantial speedup potential through parallelized modular processing. Further analysis reveals that runtime is highly correlated with the total frame count. Consequently, implementing keyframe extraction can significantly reduce the runtime without compromising the quality of the results. This is evidenced by our real-world application scenario (Fig.~\ref{fig:real2sim2real}), where using only keyframes achieved a runtime of 3,411s, significantly lower than processing full sequences from standard datasets.

\myparagraph{Real-world validation}
We further evaluate our pipeline by converting real-world scans into interactive digital twins for robotic manipulation. As shown in Fig.~\ref{fig:real2sim2real}, a mobile manipulator successfully interacts with diverse articulated objects leveraging the recovered parameters. This demonstrates that our interactive representations are sufficiently accurate and physically grounded, enabling seamless integration with downstream robotic applications.

\section{Conclusion}
We present REACT3D, a scalable zero-shot framework that transforms static 3D scenes into interactive digital twins with articulated objects. By combining open-vocabulary detection, articulation refinement, hidden-geometry completion, and seamless simulator integration, REACT3D achieves state-of-the-art performance on scene-level openable-object detection and articulation estimation. Our results demonstrate the practical utility of REACT3D in generating simulation-ready assets, providing a foundation for large-scale research in articulated scene understanding and embodied intelligence.

\bibliographystyle{IEEEtran}
\bibliography{root}

\vfill

\end{document}